%% file: main.tex
\documentclass[10pt]{iopart}

\usepackage[width=150mm,top=35mm,bottom=25mm,bindingoffset=6mm]{geometry}

\usepackage{amsmath}
\usepackage{amsfonts}

\usepackage{algorithmic}
\usepackage{algorithm}
\usepackage{array}
\usepackage[caption=false,font=normalsize,labelfont=sf,textfont=sf]{subfig}
\usepackage{textcomp}
\usepackage{stfloats}
\usepackage{url}
\usepackage{verbatim}
\usepackage{graphicx}
\usepackage{cite}
\usepackage{bm}
\usepackage[symbol]{footmisc}

\usepackage[dvipsnames]{xcolor}
\usepackage[linktocpage]{hyperref}
\hypersetup{
    colorlinks=true,
    linkcolor=BrickRed,     
    urlcolor=BrickRed,
    citecolor=BrickRed
    }

\begin{document}

\title[SymbolNet: Neural Symbolic Regression with Adaptive Dynamic Pruning for Compression]{SymbolNet: Neural Symbolic Regression with Adaptive Dynamic Pruning for Compression}

\author{Ho Fung Tsoi$^{1}$, Vladimir Loncar$^{2,3}$, Sridhara Dasu$^{1}$, Philip Harris$^{2,4}$}

\address{$^{1}$University of Wisconsin-Madison, USA \\
$^{2}$Massachusetts Institute of Technology, USA \\
$^{3}$Institute of Physics Belgrade, Serbia \\
$^{4}$Institute for Artificial Intelligence and Fundamental Interactions, USA}

\ead{ho.fung.tsoi@cern.ch}
\vspace{10pt}

%
\vspace{2pc}
\noindent{\it Keywords}: Symbolic Regression, Neural Network, Dynamic Pruning, Model Compression, Low Latency, FPGA
%
%
%
%

\input{0_abstract}

\clearpage

\tableofcontents

\input{1_intro}

\input{2_related}

\input{3_method}

\input{4_exp}

\input{5_results}

\input{6_conclusion}

\section*{Acknowledgements}
H.F.T. and S.D. are supported by the U.S. Department of Energy under the contract DE-SC0017647.
V.L. and P.H. are supported by the NSF Institute for Accelerated AI Algorithms for Data-Driven Discovery (A3D3), under the NSF grant \#PHY-2117997.
P.H. is also supported by the Institute for Artificial Intelligence and Fundamental Interactions (IAIFI), under the NSF grant \#PHY-2019786.

\section*{References}
\bibliographystyle{naturemag}
\bibliography{references}

\end{document}

%% file: 0_abstract.tex
\begin{abstract}
\normalsize
Compact symbolic expressions have been shown to be more efficient than neural network models in terms of resource consumption and inference speed when implemented on custom hardware such as FPGAs, while maintaining comparable accuracy~\cite{tsoi2023symbolic}.
These capabilities are highly valuable in environments with stringent computational resource constraints, such as high-energy physics experiments at the CERN Large Hadron Collider.
However, finding compact expressions for high-dimensional datasets remains challenging due to the inherent limitations of genetic programming, the search algorithm of most symbolic regression methods.
Contrary to genetic programming, the neural network approach to symbolic regression offers scalability to high-dimensional inputs and leverages gradient methods for faster equation searching.
Common ways of constraining expression complexity often involve multistage pruning with fine-tuning, which can result in significant performance loss.
In this work, we propose $\tt{SymbolNet}$, a neural network approach to symbolic regression specifically designed as a model compression technique, aimed at enabling low-latency inference for high-dimensional inputs on custom hardware such as FPGAs.
This framework allows dynamic pruning of model weights, input features, and mathematical operators in a single training process, where both training loss and expression complexity are optimized simultaneously.
We introduce a sparsity regularization term for each pruning type, which can adaptively adjust its strength, leading to convergence at a target sparsity ratio.
Unlike most existing symbolic regression methods that struggle with datasets containing more than $\mathcal{O}(10)$ inputs, we demonstrate the effectiveness of our model on the LHC jet tagging task (16 inputs), MNIST (784 inputs), and SVHN (3072 inputs).
\end{abstract}

%% file: 1_intro.tex
\section{Introduction}
Symbolic regression (SR) is a supervised learning method that searches for analytic expressions that best fit the data (see Ref.\cite{lacava2021contemporarysymbolicregressionmethods} for recent efforts).
Unlike traditional regression methods, such as linear and polynomial regression, SR can model a much broader range of complex datasets because it does not require a pre-defined functional form, which itself is dynamically evolving in the fit.

By expressing models in symbolic forms, SR facilitates human interpretation of the data, enabling the potential inference of underlying principles governing the observed system, in contrast to the opaque nature of black-box deep learning (DL) models.
A historical example is Max Planck's 1900 empirical fitting of a formula to the black-body radiation spectrum \cite{Planck1900}, known as Planck's law.
This symbolically fitted function not only inspired the physical derivation of the law but also played a key role in the revolutionary development of quantum theory.
Moreover, due to its compact representation compared to most DL models, SR can also be used as a distillation method for model compression~\cite{tsoi2023symbolic}.
This can accelerate inference time and reduce computational costs, making it particularly valuable in resource-constrained environments.
However, a significant drawback of SR is its inherent complexity.
The search space for equations expands exponentially with the number of building blocks (variables, mathematical operators, and constants), making it a challenging combinatorial problem.
In fact, finding the optimal candidate has been shown to be NP-hard \cite{virgolin2022symbolic}.

Genetic programming (GP) has traditionally been the primary approach to SR~\cite{Schmidt2009,cranmer2023interpretable,gplearn,Operon,Virgolin_2021}.
It constructs expressions using a tree representation, where the algebraic relations are reflected in the tree's structure.
The tree's lowest nodes consist of constants and variables, while the nodes above represent mathematical operations.
GP grows an expression tree in a manner that mimics biological evolution, employing node mutations and subtree crossovers to explore variations in expressions.
Candidates are grouped into generations and participate in a tournament selection process, where individuals with the highest fitness scores survive and advance.
Although GP has been successful in discovering human-interpretable solutions for many low-dimensional problems, its discrete combinatorial approach and lengthy search times make it unsuitable for large and high-dimensional datasets.

An alternative approach to SR involves using a DL framework~\cite{eql2016,eql2018,eql2021,Kim_2021,abdellaoui2021symbolic,costa2021fast,petersen2021deep,zhou2022bayesian,Kubal_k_2023}, such as training a neural network (NN) with activation functions that generalize to broader mathematical operations, including unary functions like $(\cdot$)$^2$ and sin($\cdot$), as well as binary functions like $+$ and $\times$.
The NN is trained with enforced sparse connections, ensuring that the final expressions derived from the NN are compact enough to be human-interpretable or efficiently deployable in resource-constrained environments.
In addition to benefiting from faster gradient-based optimization, the DL approach can utilize GPUs to accelerate both training and inference, whereas GP-based algorithms are typically limited to CPUs.
The key to training a NN that produces effective SR results, balancing model performance and complexity, is controlling sparsity.
However, most recent developments in using NNs for SR have relied on less efficient pruning methods to achieve the necessary sparsity, requiring multiple training phases with hard-threshold pruning followed by fine-tuning.
These multistage frameworks often result in significant performance compromise, as accuracy and sparsity are optimized in separate training phases.
The lack of an integrated sparsity control scheme prevents DL approaches from fully realizing their potential to expand SR's applicability to a broader range of problems.

In this contribution, we introduce $\tt{SymbolNet}$\footnote[2]{A tensorflow \cite{tensorflow2015-whitepaper} implementation code is available at \url{https://github.com/hftsoi/SymbolNet}}, a DL approach to SR utilizing NN in a novel and SR-dedicated pruning framework, specifically designed as a model compression technique for low-latency inference applications, with the following properties:
\begin{itemize}
    \item \textbf{End-to-end single-phase dynamic pruning.} It requires only a single training phase without the need for fine-tuning.
    Unlike traditional methods that rely on a pre-specified threshold for `heavy-hammer-style' pruning, this framework, inspired by dynamic sparse training \cite{liu2020dynamic}, introduces a trainable threshold associated with each model weight. The pruning of a weight is automatically determined by the dynamic competition between the weight and its threshold.
    We extend this concept to also prune input features, automating feature selection within the framework without the need for external packages or additional steps.
    Similarly, we introduce operator pruning, which involves dynamically transforming more complex mathematical operators into simpler arithmetic operations.
    Overall, a trainable threshold is independently assigned to each model weight, input feature, and mathematical operator, enabling dynamic pruning to occur within a single training phase.
    \item \textbf{Convergence to the desired sparsity ratios.} For each of the four pruning types---model weights, input features, unary operators, binary operators---we introduce a regularization term that adaptively adjusts its strength in relation to the training loss. This allows the model to converge to the desired sparsity ratios, as specified by the user.
    \item \textbf{Scalability of SR to high-dimensional datasets.} Dynamic pruning can enforce strong sparsity while being optimized simultaneously with model performance. Combined with gradient-based optimization, this approach enables the generation of optimal and compact expressions that can effectively fit large and complex datasets.
\end{itemize}

As far as we are aware, most of the SR literature has primarily tested their methods on datasets with input dimensions below $\mathcal{O}(10)$.
These methods have yet to be demonstrated as efficient solutions for high-dimensional problems such as MNIST and beyond.
Our framework leverages the compact representation of symbolic expressions and is designed to enable low-latency inference on custom hardware such as FPGAs, targeting high-dimensional datasets without ground-truth equations, as commonly encountered in high-energy experiments at the CERN Large Hadron Collider (LHC).
We validate our framework by learning compact expressions from datasets with input dimensions ranging from $\mathcal{O}(10)$ to $\mathcal{O}(1000)$, demonstrating the effectiveness of the model in solving more practical problems for deployment with constrained computational resources.

The paper is structured as follows.
Sec.~\ref{sec-related-work} discusses some of the previous efforts related to this work.
Sec.~\ref{sec:method} details the architecture and training framework of $\tt{SymbolNet}$.
Sec.~\ref{sec-exp} describes the datasets and outlines the experiments.
Sec.~\ref{sec-results} presents the results, comparing $\tt{SymbolNet}$ with baseline methods in obtaining compact and competitive expressions, and also comparing $\tt{SymbolNet}$ with typical compression methods in the context of FPGA deployment for sub-microsecond latency.

%% file: 2_related.tex
\section{Related work}
\label{sec-related-work}

SR has started to gain significant attention over the past decade~\cite{doi:10.1126/sciadv.aay2631,Keren_2023,Kaptanoglu_2022,lacava2021contemporarysymbolicregressionmethods}.
Approaches to SR have traditionally been based on GP, first formulated in \cite{Koza1994}, arising from the idea of creating a program that enables a computer to solve problems in a manner similar to natural selection and genetic evolution.
$\tt{Eureqa}$ \cite{Schmidt2009} is one of the first GP-based SR libraries, but it was developed as a commercial proprietary tool, limiting its accessibility for scientific research.
$\tt{PySR}$ \cite{cranmer2023interpretable} is a recently developed open-source library built upon the classic GP approach, augmented with a novel evolve-simplify-optimize loop, making it suitable for practical SR and the automatic discovery of scientific equations~\cite{wadekar2020modeling,Shao_2022,Delgado_2022,Wadekar_2023_b,lemos2022rediscovering,Wadekar_2023_a,grundner2023datadriven}.
Other examples include $\tt{gplearn}$ \cite{gplearn}, $\tt{Operon}$ \cite{Operon}, and $\tt{GP-GOMEA}$ \cite{Virgolin_2021}.
For scalable SR, feature selection can be employed prior to performing the equation search.
For instance, $\tt{PySR}$ integrates an external random forest regression algorithm to handle high-dimensional datasets by selecting a subset of features based on their relevance to a regression task, which are then passed to the main GP algorithm.
However, this approach is practically viable only if the number of selected features remains low, such as below ten, as the search algorithm is still being performed with GP.
Alternative methods include the fast function extraction~\cite{McConaghy2011,Kammerer_2022} and differentiable GP approach~\cite{zeng2023differentiablegeneticprogramminghighdimensional}, which aim to address the scalability challenges of SR but are not NN-based.

DL has been highly successful in addressing complex problems in fields such as computer vision and natural language processing~\cite{deeplearning}, yet, its application in the domain of SR has not been thoroughly investigated.
Equation learner ($\tt{EQL}$) \cite {eql2016,eql2018,eql2021} is one of the first NN architectures proposed for performing SR. The approach involves constructing a NN using primitive mathematical operations for neuron activation, training it to achieve a sparse structure through pruning~\cite{NIPS1989_6c9882bb,louizos2018learningsparseneuralnetworks,han2016deepcompressioncompressingdeep}, and then unrolling it to obtain the final expression.
To prevent the formation of overly complex equations, a three-stage training scheme is employed.
In the first stage, a fully-connected NN is trained without regularization, focusing solely on minimizing regression error to allow the parameters to vary freely and establish a solid starting point.
In the second stage, $L_{1}$ regularization is imposed to encourage small weights, leading to the emergence of a sparse connection pattern.
Finally, all weights below a certain threshold are set to zero and frozen, effectively enforcing a fixed $L_{0}$ norm.
The remaining weights are fine-tuned without any regularization.
Later research demonstrated that using the $L_{0.5}^{*}$ regularizer \cite{Kim_2021,abdellaoui2021symbolic}, which is constructed from a piecewise function and serves as a smooth variant of $L_{0.5}$, can enforce stronger sparsity than $L_{1}$.
These methods were primarily tested on simple dynamic system problems with input dimensions of $\mathcal{O}(1)$.

Other variants of NN-based approaches to SR include $\tt{OccamNet}$ \cite{costa2021fast}, which uses a NN to define a probability distribution over a function space, optimized using a two-step method that first samples functions and then updates the weights so that the probability mass is more likely to produce better-fitting functions.
Sparsity is maximized by introducing temperature-controlled softmax layers that sample sparse paths through the NN.
$\tt{DSR}$ \cite{petersen2021deep} employs an autoregressive recurrent NN to generate expressions sequentially, optimizing them based on reinforcement learning.
$\tt{MathONet}$ \cite{zhou2022bayesian} is a Bayesian learning framework that incorporates sparsity as priors and is applied to solve differential equations.
$\tt{N4SR}$ \cite{Kubal_k_2023} is a multistage learning framework that allows integration of domain-specific prior knowledge.
Another class of approaches utilizes transformers pre-trained on large-scale synthetic datasets to generate symbolic expressions from data.
For those interested, further details can be found in \cite{pmlr-v139-biggio21a,valipour2021symbolicgpt,kamienny2022endtoend,vastl2022symformer}.

However, these approaches primarily focus on small and low-dimensional datasets and do not explore scalability.
GP-based methods are inherently difficult to scale due to their discrete search strategies, while DL-based methods lack an efficient and SR-dedicated approach to constrain model size.
Our method, detailed in the following section, attempts to address this gap.

%% file: 3_method.tex
\section{SymbolNet architecture}
\label{sec:method}

In this section, we describe the model architecture and training framework.
We construct a NN composed of symbolic layers, using generic mathematical operators as activation functions.
Trainable thresholds are introduced to dynamically prune model weights, input features, and operators.
Additionally, a self-adaptive regularization term is introduced for each pruning type to ensure convergence to the target sparsity ratio.

\subsection{Neural symbolic regression}
\label{sec:Neural symbolic regression}

\begin{figure}[!t]
\centering
\includegraphics[width=0.3\textwidth]{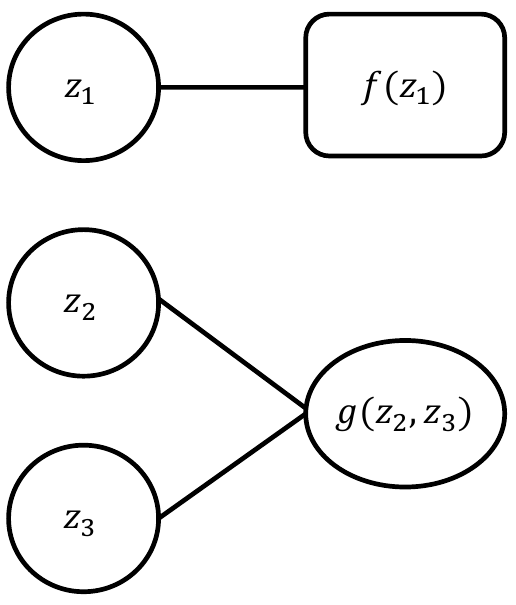}
\caption{A symbolic layer composed of three linear transformation nodes $z$, activated by a unary operator $f$ and a binary operator $g$.}
\label{fig:arch0}
\end{figure}

\begin{figure}[!t]
\centering
\includegraphics[width=1\textwidth]{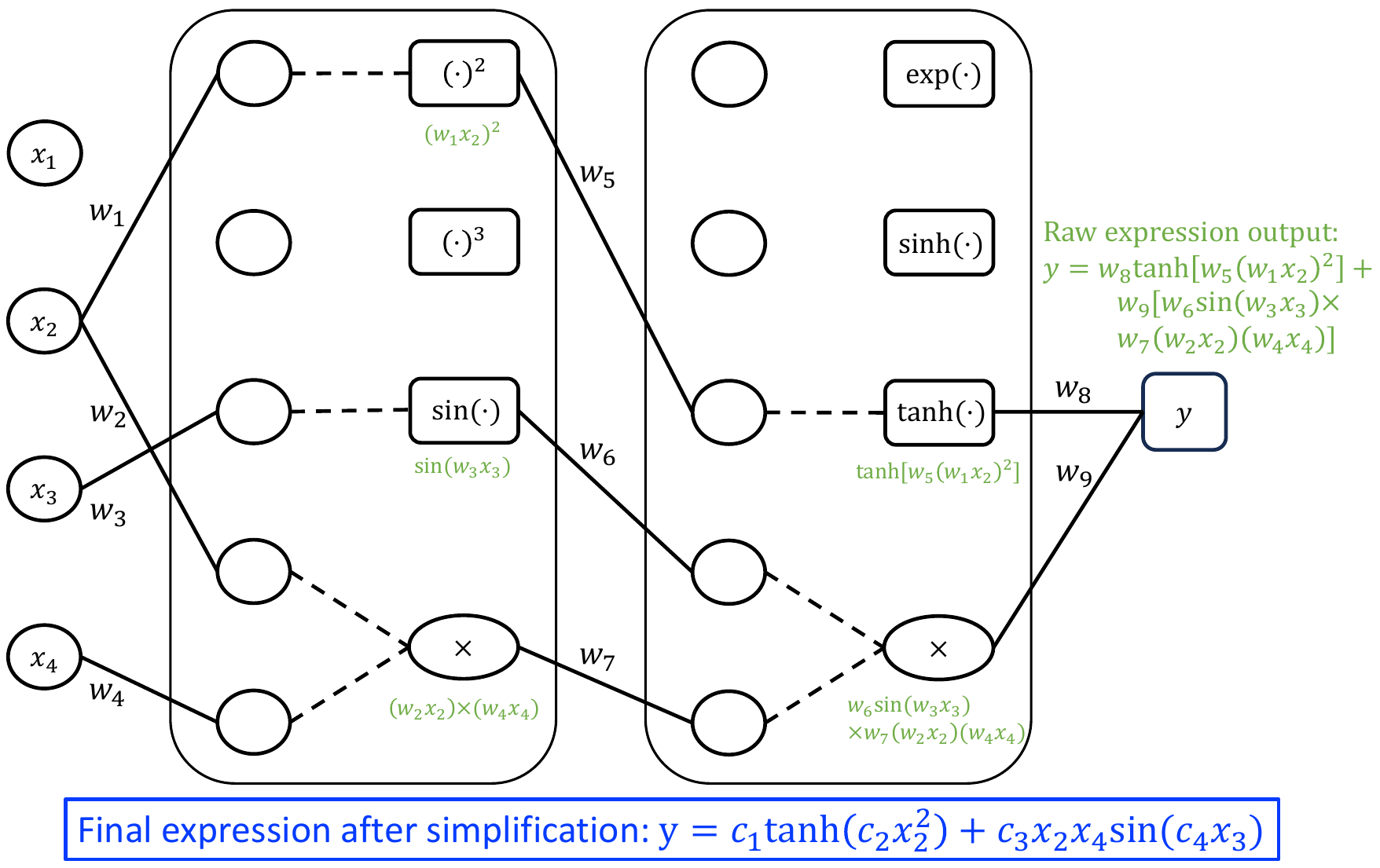}
\caption{An example NN with four input features ($x$), two symbolic layers (large rectangles), and one output node ($y$). Each symbolic layer contains five linear transformations (empty circles), followed by three unary operations (small rectangles) and one binary operation (ovals). The solid lines represent nonzero model weights ($w$) for the linear transformation, while the dashed lines indicate activation by mathematical operations. The intermediate expression outputs are shown as green text. The final expression from this example model, after simplifying the constants from $w$ to $c$, is shown in blue text: $\bm{y=c_1\tanh(c_2x_2^2)+c_3x_2x_4\sin(c_4x_3)}$. This illustrates the basic architecture of $\tt{SymbolNet}$ before incorporating additional components for adaptive dynamic pruning.}
\label{fig:arch1}
\end{figure}

We adapt the $\tt{EQL}$ architecture introduced in \cite{eql2016} as our starting point for approaching SR using NNs.
This basic architecture functions similarly to a multilayer perceptron~\cite{Rosenblatt1958ThePA,Cybenko1989ApproximationBS,Hornik1989MultilayerFN}, with the key difference being that each hidden layer is generalized to a symbolic layer.
A symbolic layer consists of two operations: the standard linear transformation of outputs from the previous layer, followed by a layer of heterogeneous unary ($f(x)$: $\mathbb{R}\rightarrow\mathbb{R}$, e.g., $x^{2}$, $\sin(x)$ and $\exp(-(x)^{2})$) and binary ($g(x,y)$: $\mathbb{R}^{2}\rightarrow\mathbb{R}$, e.g., $x+y$, $xy$, and $\sin(x)\cos(y)$) operations as activation functions.
Thus, if a symbolic layer contains $u$ unary operators and $b$ binary operators, its input dimension is $u+2b$ and its output dimension is $u+b$.
An example of a symbolic layer is illustrated in Fig.~\ref{fig:arch0}, where $u=1$ and $b=1$.

\subsection{Dynamic pruning per network component type}
\label{sec:Dynamic pruning}

The expressiveness of NNs is partly due to the large number of adjustable parameters they contain.
Even a shallow NN can be over-parameterized in the context of symbolic representation.
Therefore, controlling sparsity is crucial for NN-based SR, while still maintaining reasonable model performance.
An example of a sparsely connected NN composed of symbolic layers, which generates a compact expression, is illustrated in Fig. \ref{fig:arch1}.

Dynamic sparse training, introduced in \cite{liu2020dynamic}, is an improved alternative to the traditional `heavy-hammer-style' pruning that applies a fixed threshold for all model weights.
Instead, this method defines a threshold vector for each layer, where the thresholds are trainable and can be updated through backpropagation~\cite{Rumelhart1986LearningRB}.
This makes pruning a dynamic process, with weights and their associated thresholds continuously competing.
Pruning occurs more precisely as it takes place at each training step rather than between epochs.
Since both the weight and its threshold continue to update even after pruning, a pruned weight can potentially be recovered if the competition reverses.
This approach follows a single training schedule, eliminating the need for multistage training with separate fine-tuning, while simultaneously optimizing both model weights and sparsity ratios.
This simple yet effective framework is particularly valuable for NN-based SR.

Inspired by this method, we incorporate it into our model to perform a weight-wise pruning.
We then generalize the idea to also prune input features and mathematical operators, introducing a regularization strategy to achieve convergence to the desired sparsity ratios.

To implement dynamic pruning, Ref. \cite{liu2020dynamic} used a step function as a masking function, along with a piecewise polynomial estimator that is nonzero and finite in the range $[-1,1]$ but zero elsewhere.
This approach was used to approximate the derivative in the backward pass, preventing a zero gradient in the thresholds, since the derivative of the original step function is almost zero everywhere.
In this work, we employ a similar binary step function $\theta(x)=\mathbf{1}_{x>0}$ (i.e., 1 if $x>0$, otherwise 0) for masking in the forward pass, along with a smoother estimator in the backward pass using the derivative of the sigmoid function $\frac{d\theta(x)}{dx}\approx\frac{\kappa e^{-\kappa x}}{(1+e^{-\kappa x})^2}$ with $\kappa=5$ (a very high $\kappa$ would effectively turn it into a delta function, while a very low $\kappa$ would make it flat and less sensitive to the step location), which is nonzero everywhere.

\begin{figure*}[!t]
\centering
\subfloat[]{\includegraphics[width=0.5\textwidth]{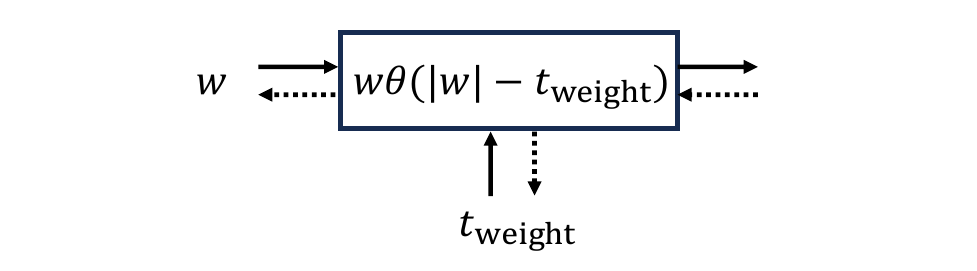}%
\label{fig:pruning_weight}}
\hfil
\subfloat[]{\includegraphics[width=0.5\textwidth]{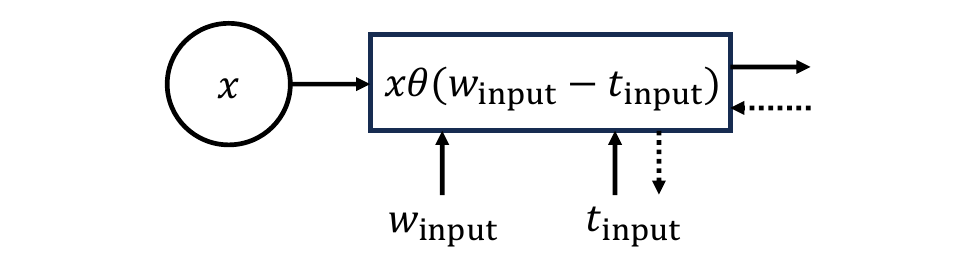}%
\label{fig:pruning_input}}\\
\subfloat[]{\includegraphics[width=0.5\textwidth]{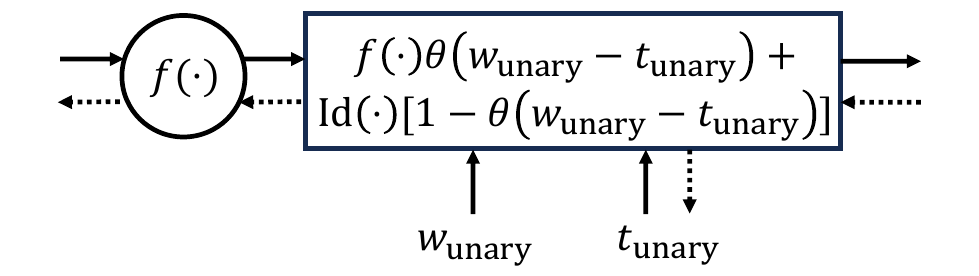}%
\label{fig:pruning_unary}}
\hfil
\subfloat[]{\includegraphics[width=0.5\textwidth]{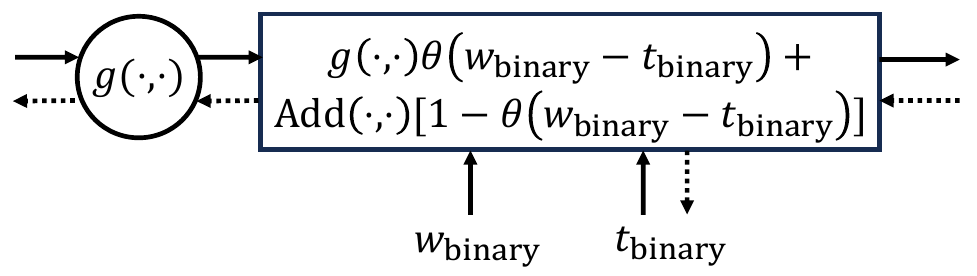}%
\label{fig:pruning_binary}}
\caption{Schematic sketch of the SR-dedicated dynamic pruning mechanism within the $\tt{SymbolNet}$ architecture: (a) model weights, (b) input features, (c) unary operators, and (d) binary operators. Solid arrows represent the forward pass, while dotted arrows represent the backward pass, linking the trainable parameters. The $\tt{SymbolNet}$ architecture is constructed by integrating these elements with the basic network architecture illustrated, for example, in Fig. \ref{fig:arch1}.}
\label{fig:pruning}
\end{figure*}

We introduce a pruning mechanism for model weights, input features, unary operators, and binary operators, as illustrated in Fig.~\ref{fig:pruning} and explained in the following subsections.

\subsubsection{\bf{Pruning of model weights}}
\label{sec:Sparse weights}

For each model weight $w$, we associate a trainable pruning threshold $t_{\text{weight}}$.

In each layer that performs a linear transformation with input dimension $n$ and output dimension $m$, there are a weight matrix and a threshold matrix: $\bm{w},\bm{t}_{\text{weight}}\in\mathbb{R}^{n\times m+m}$, where the second $m$ corresponds to the number of bias terms.
Each threshold is initialized to zero and is clipped to the range of $[0,\infty)$ during parameter updates, as the magnitude of each $w$ is unbounded.

The weight-wise pruning in each layer is implemented using the step function matrix $\bm{\Theta}_{\text{weight}}=\theta(|\bm{w}|-\bm{t}_{\text{weight}})\in\mathbb{R}^{n\times m+m}$, applied as follows: $w_{ij}\rightarrow(\bm{w}\odot \bm{\Theta}_{\text{weight}})_{ij}=w_{ij}\theta(|w_{ij}|-t_{\text{weight},ij})$, as illustrated in Fig.~\ref{fig:pruning_weight}.

Concatenating all layers of the architecture, we denote $\bm{W}=\{\bm{w}\}$ and $\bm{T}_{\text{weight}}=\{\bm{t}_{\text{weight}}\}$, with the total dimension being the total number of weights $n_{\text{weight}}$.

\subsubsection{\bf{Pruning of input features}}
\label{sec:Sparse inputs}

Similarly, for each input node $x$, we associate it with an auxiliary weight $w_{\text{input}}$ and a trainable pruning threshold $t_{\text{input}}$.

Since there is already a linear transformation of the inputs when propagating to the next layer, and because only the relative distance between the weight and threshold matters on the pruning side, a trainable auxiliary weight would be redundant.
Therefore, we fix all $w_{\text{input}}=1$ and set them as untrainable.
Each threshold is initialized to zero and is clipped to the range of $[0,1]$ during parameter updates.

For the input vector $\bm{x}\in\mathbb{R}^{n_{\text{input}}}$, there is an auxiliary weight vector and a threshold vector $\bm{W}_{\text{input}},\bm{T}_{\text{input}}\in\mathbb{R}^{n_{\text{input}}}$.
The pruning operation is implemented using the step function vector $\bm{\Theta}_{\text{input}}=\theta(\bm{W}_{\text{input}}-\bm{T}_{\text{input}})$, applied as follows: $x_{i}\rightarrow(\bm{x}\odot \bm{\Theta_{\text{input}}})_{i}=x_{i}\theta(1-T_{\text{input},i})$, as illustrated in Fig.~\ref{fig:pruning_input}.
In other words, an input node is pruned when its associated threshold value reaches 1.

\subsubsection{\bf{Pruning of mathematical operators}}
\label{sec:Sparse operators}

Instead of pruning mathematical operators to zero directly, we simplify them to basic arithmetic operations, as pruning to zero can be accomplished by pruning the corresponding weights in their linear transformations when propagating to the next layer.

Similar to input pruning, for each unary operator $f(\cdot)$: $\mathbb{R}\rightarrow\mathbb{R}$, we associate it with an untrainable auxiliary weight fixed at $w_{\text{unary}}=1$ and a trainable pruning threshold $t_{\text{unary}}\in[0,1]$.
The pruning operation is applied to each unary operator node as $f(\cdot)\rightarrow[f(\cdot)\theta(1-t_{\text{unary}})+\text{Id}(\cdot)(1-\theta(1-t_{\text{unary}}))]$, where $\text{Id}(\cdot)$ is the identity function, as illustrated in Fig.~\ref{fig:pruning_unary}.
This means a unary operator will be simplified to an identity operator when necessary, helping to prevent overfitting and reducing overly complex components, such as function nesting (e.g., $\sin(\sin(\cdot))$).
We denote all thresholds of this type by a vector $\bm{T}_{\text{unary}}$, with a dimension of $n_{\text{unary}}$, representing the total number of unary operators in the architecture.

Similarly, for each binary operator $g(\cdot,\cdot)$: $\mathbb{R}^{2}\rightarrow\mathbb{R}$, we associate it with an untrainable auxiliary weight fixed at $w_{\text{binary}}=1$ and a trainable pruning threshold $t_{\text{binary}}\in[0,1]$.
The pruning operation is applied to each binary operator node as $g(\cdot,\cdot)\rightarrow[g(\cdot,\cdot)\theta(1-t_{\text{binary}})+\text{Add}(\cdot,\cdot)(1-\theta(1-t_{\text{binary}}))]$, where $\text{Add}(\cdot,\cdot)$ is the addition operator, as illustrated in Fig.~\ref{fig:pruning_binary}.
This means a binary operator will be simplified to an addition operator when necessary, which is motivated by the fact that addition is typically simpler to compute than other more complex binary operators.
For instance, an addition operation requires only one clock cycle to execute on an FPGA, while other binary operators may generally take significantly longer.
We denote all thresholds of this type by a vector $\bm{T}_{\text{binary}}$, with a dimension of $n_{\text{binary}}$, representing the total number of binary operators in the architecture.

\subsection{Self-adaptive regularization for sparsity}
\label{sec:Regularization term for sparsity}

We introduce a regularization term for each threshold type to encourage large threshold values.
For the model weight thresholds, we adapt the approach in \cite{liu2020dynamic} and use the form $L_{\text{threshold}}^{\text{weight}}=\frac{1}{n_{\text{weight}}}\sum_{i=1}^{n_{\text{weight}}}\exp(-T_{\text{weight},i})$, where the sum runs over all the weight thresholds in the architecture.
Since each threshold $T_{\text{weight},i}\in[0,\infty)$ is unbounded, this sum of the exponentials will not drop to zero immediately, even if some thresholds become large.
For other types with bounded thresholds, we use a more efficient form with a single exponential: $L_{\text{threshold}}^{\text{aux}=\text{\{input,unary,binary\}}}=\exp(-\frac{1}{n_{\text{aux}}}\sum_{i=1}^{n_{\text{aux}}}T_{\text{aux},i})$.

\begin{figure}[!t]
\centering
\includegraphics[width=0.6\textwidth]{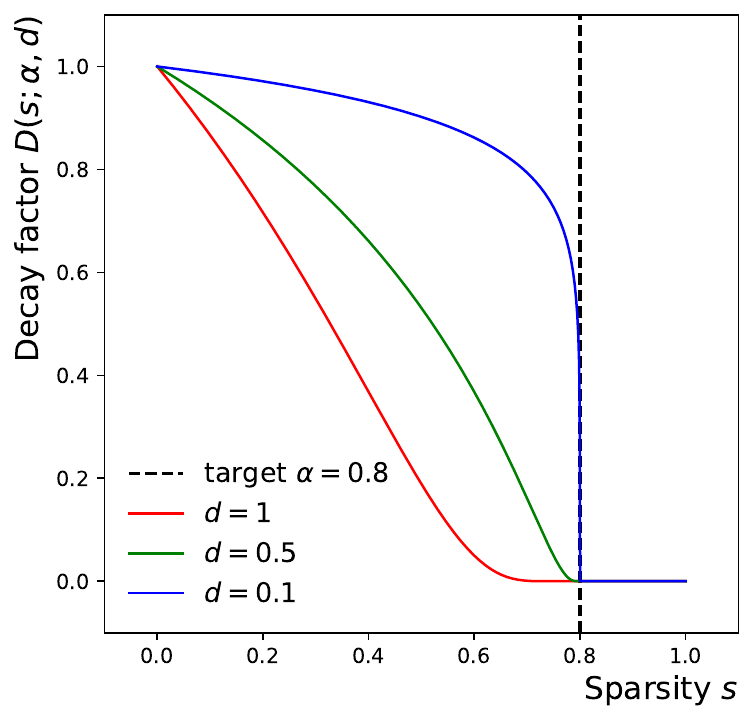}
\caption{The decay factor, $D(s;\alpha,d)$, is employed to reduce the rate of increase in high-thredhold values as the sparsity ratio ($s$) approaches its target value ($\alpha$). High-threshold driving is paused when $s\geq\alpha$. The profiles of $D(0\leq s\leq 1)$ for a target sparsity ratio of $\alpha=0.8$ at three different decay rates ($d$) are shown.}
\label{fig:decay}
\end{figure}

For each of the regularizers, we introduce a $\textit{decay factor}$:
\begin{equation}
\label{eq:decay}
D(s;\alpha,d)=\exp\bigg[-\bigg(\frac{\alpha}{\alpha-\text{min}(s,\alpha)}\bigg)^{d}+1\bigg].
\end{equation}
Here, the sparsity $s\in[0,1]$ represents the ratio of pruned parameters or operators, evaluated at each training step.
The parameter $\alpha\in[0,1]$ sets the target sparsity ratio, and $d$ is the decay rate that controls how quickly the sparsity driving slows down.
When the regularizer is multiplied by this term, it effectively slows down high-threshold driving as the sparsity ratio $s$ grows, and eliminates the regularizer when $s\geq\alpha$.
The decay factor is illustrated in Fig.~\ref{fig:decay} for different decay rates $d$.

The full regularization term for each threshold type takes the form $L_{\text{sparse}}=L_{\text{error}}D(s;\alpha,d)L_{\text{threshold}}$, where $L_{\text{error}}$ represents the training loss (e.g., regression error). 
The strength of the threshold regularizer $L_{\text{threshold}}$ is adaptively adjusted by the product $L_{\text{error}}D(s;\alpha,d)$, which is initially set by the training loss.

At the start of the training, when the sparsity ratio is initialized at $s=0$, the regularization is as significant as the training loss itself: $L_{\text{sparse}}=L_{\text{error}}$, since $L_{\text{threshold}}=D=1$.
The presence of $L_{\text{threshold}}$ drives the thresholds to increase through backpropagation, which in turn reduces $L_{\text{threshold}}$.

As the sparsity ratio begins to increase $s>0$, the decay factor $D$ drops below 1, slowing the growth of the thresholds.
This process continues until the target sparsity ratio is reached, where $D$ eventually drops to 0, or until further increasing the sparsity would significantly elevate the training loss, even if the target is not yet reached.
Thus, the regularizer $L_{\text{sparse}}$ is designed to guide the sparsity ratio toward convergence at the desired target value.

\subsection{Training framework}
\label{sec:Training framework}

Integrating all together, for a dataset $\{(\bm{x}^{i},\bm{y}^{i})\}_{i=1}^{N}$ with inputs $\bm{x}^{i}\in\mathbb{R}^{n_{\text{input}}}$ and labels $\bm{y}^{i}\in\mathbb{R}^{n_{\text{output}}}$, the algorithm aims to solve the following multi-objective optimization problem for a network $\phi:\mathbb{R}^{n_{\text{input}}}\rightarrow\mathbb{R}^{n_{\text{output}}}$ with outputs $\hat{\bm{y}}^{i}=\phi(\bm{x}^{i})$:
\begin{equation}
\label{eq:loss1}
\bm{W}^{*}, \bm{T}^{*}_{\text{weight}}, \bm{T}^{*}_{\text{aux}}=\operatorname*{arg\,min}_{\bm{W}, \bm{T}_{\text{weight}}, \bm{T}_{\text{aux}}} \mathcal{L}(\bm{W}, \bm{T}_{\text{weight}}, \bm{T}_{\text{aux}}; \alpha_{\text{weight}}, \alpha_{\text{aux}}, d),\\
\end{equation}
where
\begin{equation}
\begin{aligned}
\label{eq:loss2}
\mathcal{L} &= L_{\text{error}} + L_{\text{sparse}}^{\text{weight}} + L_{\text{sparse}}^{\text{aux}}, \\
L_{\text{sparse}}^{\text{weight}} &= L_{\text{error}}\times D(s_{\text{weight}};\alpha_{\text{weight}},d)\times\frac{1}{n_{\text{weight}}}\sum_{i=1}^{n_{\text{weight}}}\exp(-T_{\text{weight},i}),\\
L_{\text{sparse}}^{\text{aux}} &= L_{\text{error}}\times D(s_{\text{aux}};\alpha_{\text{aux}},d)\times\exp\bigg(-\frac{1}{n_{\text{aux}}}\sum_{i=1}^{n_{\text{aux}}}T_{\text{aux},i}\bigg),
\end{aligned}
\end{equation}
with $\text{aux}=\text{\{input,\,unary,\,binary\}}$.
We use the mean squared error (MSE) as the training loss: $L_{\text{MSE}} = \frac{1}{Nn_{\text{output}}}\sum_{i=1}^{N}\sum_{j=1}^{n_{\text{output}}} (y_{j}^{i}-\hat{y}_{j}^{i})^{2}$.
For our experiments presented in the next section, we set $d=0.01$ (a lower value is preferred to avoid too fast a decay rate where most weights are pruned at early epochs before the model learns anything from data).
The remaining free parameters are $\alpha_{\text{weight}}$, $\alpha_{\text{input}}$, $\alpha_{\text{unary}}$, and $\alpha_{\text{binary}}$, which respectively determine the target sparsity ratios for different types of pruning.

In this single-phase training framework, the overall sparse structure is divided into sparse substructures for the model weights, input features, unary operators, and binary operators, respectively.
In particular, a set of sparse input features is automatically searched without the need for an external feature selection process.
Each of these sparse substructures is dynamically determined by the competition between the corresponding weights and thresholds.
Furthermore, the sparse structure dynamically competes with the regression performance, allowing both to be optimized simultaneously.
This approach enables direct specification of target sparsity ratios for each component (weights, features, and operators), rather than requiring the indirect tuning of multiple regularization parameters whose relationships to the resulting sparsity levels are more difficult to predict.
The final symbolic expressions are derived by unrolling the trained network.

%% file: 4_exp.tex
\section{Experimental setup}
\label{sec-exp}

\subsection{Expression complexity}

\begin{figure}[!t]
\centering
\includegraphics[width=1\textwidth]{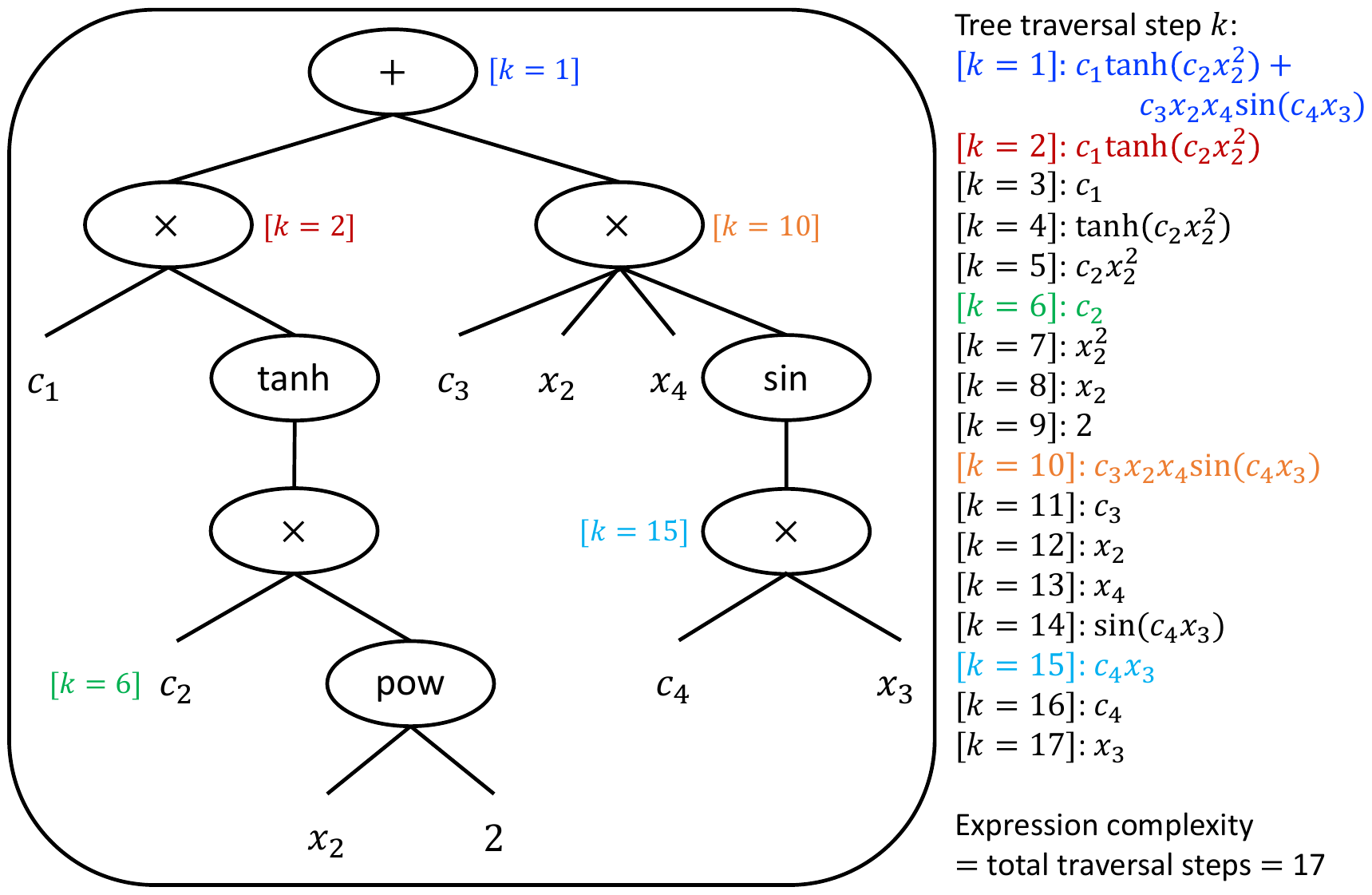}
\caption{Counting the complexity of an expression in its tree representation involves tree traversal. Using the example expression $\bm{y=c_1\tanh(c_2x_2^2)+c_3x_2x_4\sin(c_4x_3)}$ from Fig. \ref{fig:arch1}, the sub-expressions at each step of the traversal ($k$) are listed. The expression complexity of this example is 17, assuming all mathematical operators, input features, and constants are equally weighted. Note that the number of possible traversal steps corresponds to the number of nodes in the tree.}
\label{fig:arch2}
\end{figure}

To quantify the size of a symbolic model, we use a metric called expression complexity \cite{cranmer2023interpretable}.
This metric is computed by counting all possible steps involved in traversing the expression tree, which corresponds to the total number of nodes in the tree.
For example, the expression $y=c_1\tanh(c_2x_2^2)+c_3x_2x_4\sin(c_4x_3)$, generated in Fig. \ref{fig:arch1}, has a complexity of 17, as illustrated in Fig. \ref{fig:arch2}.
The $\tt{Sympy}$ library \cite{10.7717/peerj-cs.103} provides the $\textit{preorder\_traversal}$ method, which can be used to calculate the number of steps required to traverse a given expression.

We assume that all types of tree node (mathematical operator, input variable, and constant) have the same complexity of 1 during the counting process. However, this assumption may not hold universally.
For example, in the context of Field-Programmable Gate Arrays (FPGAs), computing $\tan(\cdot)$ might require significantly more clock cycles than computing $\sin(\cdot)$.
Conversely, if mathematical functions are approximated using lookup tables that each requires only one clock cycle to compute, the assumption that all unary operators are equally weighted becomes valid \cite{tsoi2023symbolic}.
The definition of complexity for each node type depends on the specific implementation and resource allocation strategy, which we do not explore in depth.
Instead, we adopt the most straightforward assumption for our experiments.

\subsection{Baseline for comparison}

From DL-based methods, we use the $\tt{EQL}$ architecture \cite{eql2016,eql2018,eql2021}, trained within a three-stage pruning framework \cite{Kim_2021}, as the baseline for comparison purposes.
In this baseline framework, model weights are regularized using the smoothed $L_{0.5}$ term, referred to as $L^{*}_{0.5}$, which is controlled by a free parameter $\lambda$:
\begin{equation}
\begin{aligned}
\mathcal{L}&=L_{\text{MSE}}+\lambda L_{0.5}^{*}\\
L_{0.5}^{*}(w)&=
  \begin{cases}
    |w|^{0.5}&,|w|\geq a\\
    \big(-\frac{w^4}{8a^3}+\frac{3w^2}{4a}+\frac{3a}{8}\big)^{0.5}&,|w|<a
  \end{cases}
  \end{aligned}
\end{equation}
where we set $a=0.01$.
In the first phase, $L_{0.5}^{*}$ is turned off ($\lambda=0$) to allow the model to fully learn the regression and establish a solid starting point for the model weights.
In the second phase, $L_{0.5}^{*}$ is activated ($\lambda>0$) to reduce the magnitudes of the weights, promoting the emergence of a sparse model structure.
After this, weights with small magnitudes are set to zero and subsequently frozen.
In the final phase, $L_{0.5}^{*}$ is turned off again ($\lambda=0$) during the fine-tuning of the sparse model.
In the experiments that follow, we configure the baseline with network sizes and operator choices similar to those used for $\tt{SymbolNet}$.
We vary $\lambda$ from $10^{-4}$ to $10^{-1}$ and adjust the hard pruning threshold from $10^{-4}$ to $10^{-1}$ to conduct a complexity scan by running multiple trials.

Additionally, we use the GP-based $\tt{PySR}$ as another baseline for comparison.
In general, the efficiency and performance of GP-based SR algorithms tend to degrade as the number of input features increases, typically beyond ten.
To address this, $\tt{PySR}$ incorporates an external feature selector based on a gradient-boosting tree algorithm to identify important features before feeding into the main search loop, particularly for datasets with high input dimensionality~\cite{cranmer2023interpretable}.
For the LHC dataset, which has 16 input features, we configures $\tt{PySR}$ to select only half of features before performing equation searches.
However, for MNIST (784 features) and SVHN (3072 features), which feature counts exceeding the LHC dataset by more than an order of magnitude, reducing the number of features to fewer than 10 makes the models less accurate, while exceeding 10 makes GP-based searches computationally impractical, especially since the outputs are multidimensional as well.
Therefore, we restrict our comparison with $\tt{PySR}$ to the LHC dataset only.

\subsection{Datasets and experiments}

We test our framework on datasets with input dimensions ranging from $\mathcal{O}(10)$ to $\mathcal{O}(1000)$: the Large Hadron Collider (LHC) jet tagging dataset with 16 inputs \cite{Pierini:LHCjet}, MNIST with 784 inputs \cite{lecun-mnisthandwrittendigit-2010}, and SVHN with 3072 inputs \cite{svhn}.
Datasets are split into train, validation, and test sets with ratios of 0.6/0.2/0.2.
Models are trained for 200 epochs with a batch size of 1024 using the $\tt{Adam}$ optimizer~\cite{kingma2017adam} and a learning rate of 0.0015.
As described in Sec.~\ref{sec:Dynamic pruning}, trainable thresholds for all pruning types are initialized to zero and are clipped to the range [0,1] when competing with the corresponding fixed threshold target of 1, except for thresholds associated with model weight pruning, which are clipped to [0,$\infty$] since the model weights are unbounded.
Model weights and biases are initialized with random normal distributions.
Typically, one to two symbolic hidden layers with $\mathcal{O}(1-10)$ unary and binary operators  are sufficient for most cases.
There is flexibility in choosing differentiable operators for activations, with elementary functions like trigonometric and exponential functions being adequate due to the extensive function space the network can represent.

\subsubsection{LHC jet tagging}

We chose a dataset from the field of high-energy physics due to the increasing demands for efficient machine learning solutions in resource-constrained environments, such as the LHC experiments \cite{Atlas:2137107,Zabi:2020gjd}.

In collider experiments, a jet refers to a cone-shaped object containing a flow of particles, which can be traced back to the decay of an unstable particle.
The process of determining the original particle from the characteristics of the jet is known as jet tagging.
The LHC jet tagging dataset consists of simulated jets produced from proton-proton collisions at the LHC and is designed to benchmark a five-class classification task: identifying a jet as originating from a light quark, gluon, W boson, Z boson, or top quark.
This classification is based on 16 physics-motivated input features constructed from detector observables, including $\big($$\sum z\,\text{log}\,z$, $C_{1}^{\beta=0,1,2}$, $C_{2}^{\beta=1,2}$, $D_{2}^{\beta=1,2}$, $D_{2}^{(\alpha,\beta)=(1,1),(1,2)}$, $M_{2}^{\beta=1,2}$, $N_{2}^{\beta=1,2}$, $m_{\text{mMDT}}$, $\text{Multiplicity}$$\big)$.
The dataset is publicly available at \cite{Pierini:LHCjet}, and further descriptions can be found in \cite{Duarte:2018ite,Moreno:2019bmu,Coleman:2017fiq}.

\begin{figure}[!t]
\centering
\subfloat[]{\includegraphics[width=0.9\textwidth]{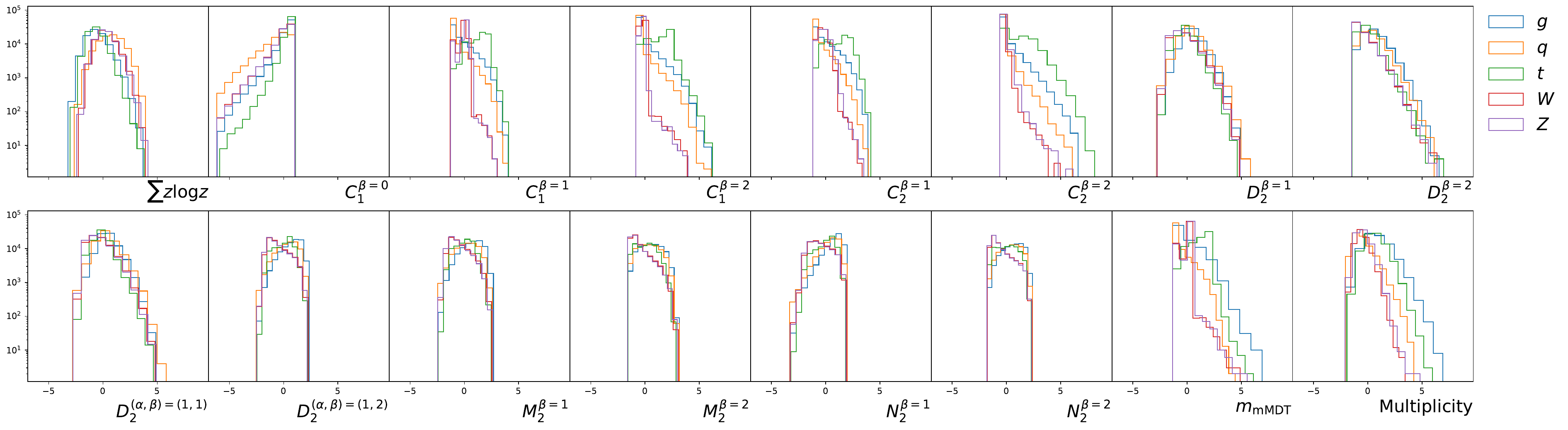}%
\label{fig:input_lhc}}
\hfil
\subfloat[]{\includegraphics[width=0.8\textwidth]{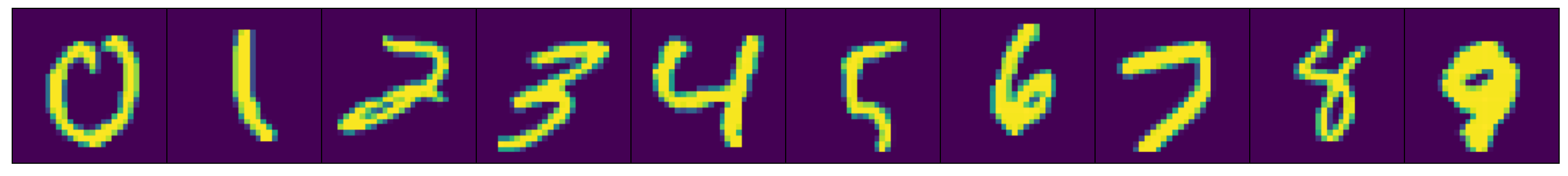}%
\label{fig:input_mnist}}
\hfil
\subfloat[]{\includegraphics[width=0.25\textwidth]{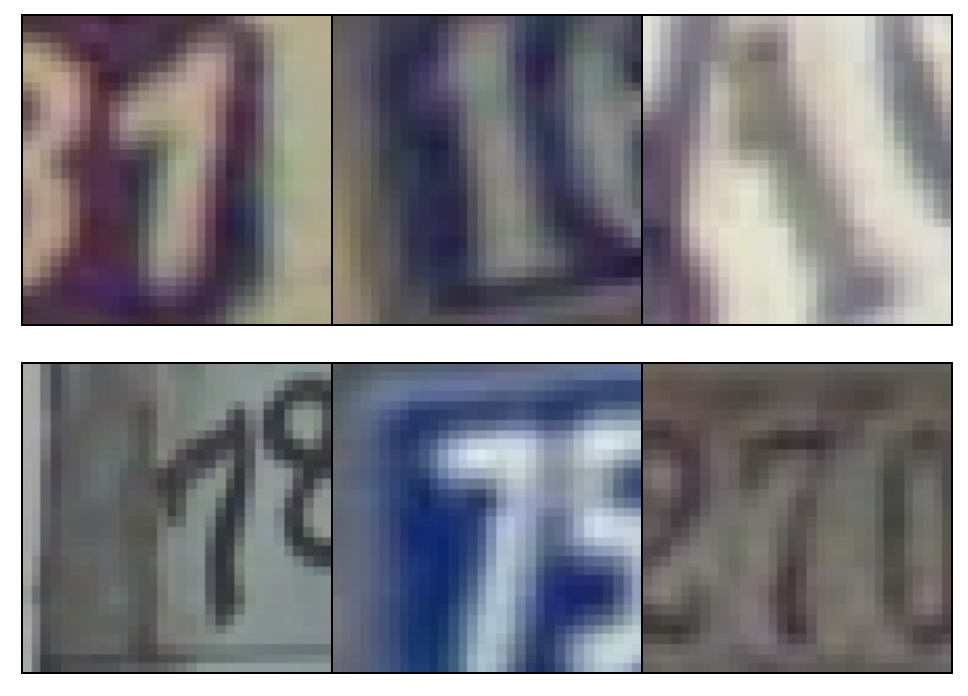}%
\label{fig:input_svhn}}
\caption{(a): Distributions of the 16 standardized input features in the LHC jet tagging dataset. From top left to top right: $\sum z\,\text{log}\,z$, $C_{1}^{\beta=0}$, $C_{1}^{\beta=1}$, $C_{1}^{\beta=2}$, $C_{2}^{\beta=1}$, $C_{2}^{\beta=2}$, $D_{2}^{\beta=1}$, and $D_{2}^{\beta=2}$. From bottom left to bottom right: $D_{2}^{(\alpha,\beta)=(1,1)}$, $D_{2}^{(\alpha,\beta)=(1,2)}$, $M_{2}^{\beta=1}$, $M_{2}^{\beta=2}$, $N_{2}^{\beta=1}$, $N_{2}^{\beta=2}$, $m_{\text{mMDT}}$, and $\text{Multiplicity}$. The five jet classes are plotted separately: gluon (blue), light quark (orange), top quark (green), W boson (red), and Z boson (purple). (b): MNIST input images with one example per class. (c): SVHN input images with three examples from the digit `1' class (top) and the digit `7' class (bottom), respectively.}
\label{fig:input_lhc-mnist-svhn}
\end{figure}

The inputs are standardized and their distributions are shown in Fig. \ref{fig:input_lhc}.
The labels are one-hot encoded.
We train $\tt{SymbolNet}$ with five output nodes, each generating an expression corresponding to one of the five jet substructure classes.
We consider models with one or two hidden symbolic layers, using unary operators including $\sin(\cdot)$, $\cos(\cdot)$, $\exp(\cdot)$, $\exp(-(\cdot)^2)$, $\sinh(\cdot)$, $\cosh(\cdot)$, and $\tanh(\cdot)$, and binary operators including $+$ and $\times$.
To explore a range of expression complexities, we conduct 40 trials for a grid search of the hyperparameters, varying the number of operators within the ranges $u\in[2,30]$ for unary operators and $b\in[2,30]$ for binary operators.
We also vary the target sparsity ratios within the ranges $\alpha_{\text{weight}}\in[0.6,0.99]$, $\alpha_{\text{input}}\in[0.4,0.9]$, $\alpha_{\text{unary}}\in[0.2,0.5]$, and $\alpha_{\text{binary}}\in[0.2,0.5]$.

\subsubsection{MNIST and binary SVHN}

We aim to demonstrate that $\tt{SymbolNet}$ can handle datasets with high input dimensions, which most existing SR methods cannot process efficiently.
To illustrate this, we examine the MNIST and SVHN datasets.
Our objective is not to create an exhaustive classifier with state-of-the-art accuracy, but rather to show $\tt{SymbolNet}$'s capability to generate simple expressions that can fit high-dimensional data with reasonable accuracy.

The MNIST dataset consists of grayscale images of handwritten digits ranging from `0' to `9', each with an input dimension of $28\times28\times1$.
The task is to classify the correct digit in a given image.
Fig.~\ref{fig:input_mnist} shows an example input image for each of the ten classes.
The inputs are flattened into a 1D array, denoted $x_{0},...,x_{783}$, and scaled to the range of $[0,1]$.
We train $\tt{SymbolNet}$ with ten output nodes, each generating an expression corresponding to one of the ten digit classes.
We consider models with one or two hidden symbolic layers, using unary operators including $\sin(\cdot)$, $\cos(\cdot)$, $\exp(-(\cdot)^2)$, and $\tanh(\cdot)$, and binary operators including $+$ and $\times$.
To explore a range of expression complexities, we conduct 40 trials, for a grid search of hyperparameters, by varying the number of operators within the ranges $u\in[2,20]$ for unary operators and $b\in[2,20]$ for binary operators.
We also vary the target sparsity ratios within the ranges $\alpha_{\text{weight}}\in[0.7,0.999]$, $\alpha_{\text{input}}\in[0.6,0.99]$, $\alpha_{\text{unary}}\in[0.2,0.5]$, and $\alpha_{\text{binary}}\in[0.2,0.5]$.

Similar to MNIST but more challenging, the SVHN dataset consists of digit images with an input dimension of $32\times32\times3$ in RGB format, where the digits are taken from noisy real-world scenes.
These images often include various distractors alongside the digit of interest, making the 10-class classification particularly challenging without additional architectural modifications, such as convolutional layers or other techniques to handle the complexity.
Therefore, for simplicity, we focus on binary classification between the digits `1' and `7'.
Fig.~\ref{fig:input_svhn} shows some example input images for each of these two classes.
The inputs are flattened into a 1D array, denoted $x_{0},...,x_{3071}$,and scaled to the range of $[0,1]$.
In this binary setting, we label the digit `1' as 0 and the digit `7' as 1, so $\tt{SymbolNet}$ has one output node and generates one expression per model.
We consider a single hidden symbolic layer, with unary operators including $\sin(\cdot)$, $\cos(\cdot)$, $\exp(-(\cdot)^2)$, and $\tanh(\cdot)$, and binary operators including $+$ and $\times$.
To explore a range of expression complexities, we conduct 40 trials, for a grid search of hyperparameters, by varying the number of operators within the ranges $u\in[2,20]$ for unary operatos and $b\in[2,10]$ for binary operators.
We also vary the target sparsity ratios within the ranges $\alpha_{\text{weight}}\in[0.8,0.999]$, $\alpha_{\text{input}}\in[0.8,0.999]$, $\alpha_{\text{unary}}\in[0.2,0.5]$, and $\alpha_{\text{binary}}\in[0.2,0.5]$.

\subsection{FPGA resource utilization and latency}
It has been demonstrated in \cite{tsoi2023symbolic} that symbolic models can potentially reduce FPGA resource consumption by orders of magnitude and achieve significantly lower latency compared to quantized yet unpruned NNs when applied to the LHC jet tagging dataset.
In this study, we perform a similar comparison, but this time between symbolic models and NNs that are strongly quantized~\cite{Coelho:2020zfu,han2016deepcompressioncompressingdeep,pmlr-v37-gupta15} and pruned~\cite{NIPS1989_6c9882bb,louizos2018learningsparseneuralnetworks,han2016deepcompressioncompressingdeep} as a typical compression strategy.

For the LHC jet tagging dataset, the baseline architecture is adopted from \cite{Duarte:2018ite}, which is a fully-connected NN, or DNN, with three hidden layers, consisting of 64, 32, and 32 neurons, respectively.
For the MNIST and SVHN datasets, the baseline architecture is adopted from \cite{Aarrestad:2021zos}, which is a convolutional NN~\cite{NIPS1989_53c3bce6,726791}, or CNN, consisting of three convolutional blocks with 16, 16, and 24 filters of size $3\times 3$, respectively, followed by a DNN with two hidden layers consisting of 42 and 64 neurons, respectively.
These baseline architectures were selected with consideration that the models are small enough to fit within the resource budget of a single FPGA board.

The baseline NNs are further compressed through quantization and pruning.
The models are trained using quantization-aware techniques with the $\tt{QKeras}$ library\footnote[3]{\url{https://github.com/google/qkeras}} \cite{Coelho:2020zfu}, and pruning is performed using the $\tt{Tensorflow}$ pruning API \cite{zhu2017prune}.
We quantize model weights and activation functions in all hidden layers to a fixed total bit width of 6, with no integer bit (denoted as $\langle 6,0\rangle$).
The model weights are pruned to achieve a sparsity ratio of approximately 90\%.

Both the symbolic models and the NNs are converted to FPGA firmware using the $\tt{hls4ml}$ library\footnote[4]{\url{https://github.com/fastmachinelearning/hls4ml}}~\cite{vloncar_2021_5680908,Duarte:2018ite,tsoi2023symbolic}.
Synthesis is performed with Vivado HLS (2020.1) \cite{vivado}, targeting a Xilinx Virtex UltraScale+ VU9P FPGA (part no.: xcvu9p-flga2577-2-e), with the clock frequency set to 200 MHz (or clock period of 5 ns).
We compare FPGA resource utilization and inference latency between symbolic models learned by $\tt{SymbolNet}$ and the compressed NNs.

%% file: 5_results.tex
\section{Results}
\label{sec-results}

\subsection{LHC jet tagging}

\begin{figure*}[!t]
\centering
\subfloat[]{\includegraphics[width=0.32\textwidth]{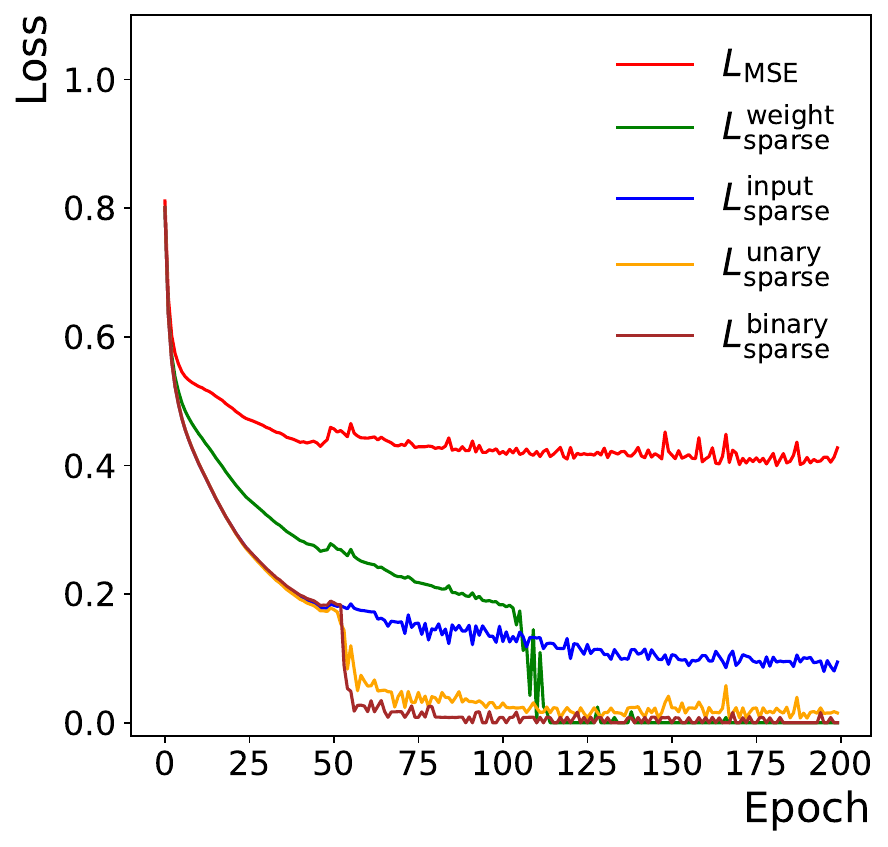}}%
\label{fig:training_lhc_1}
\hfil
\subfloat[]{\includegraphics[width=0.32\textwidth]{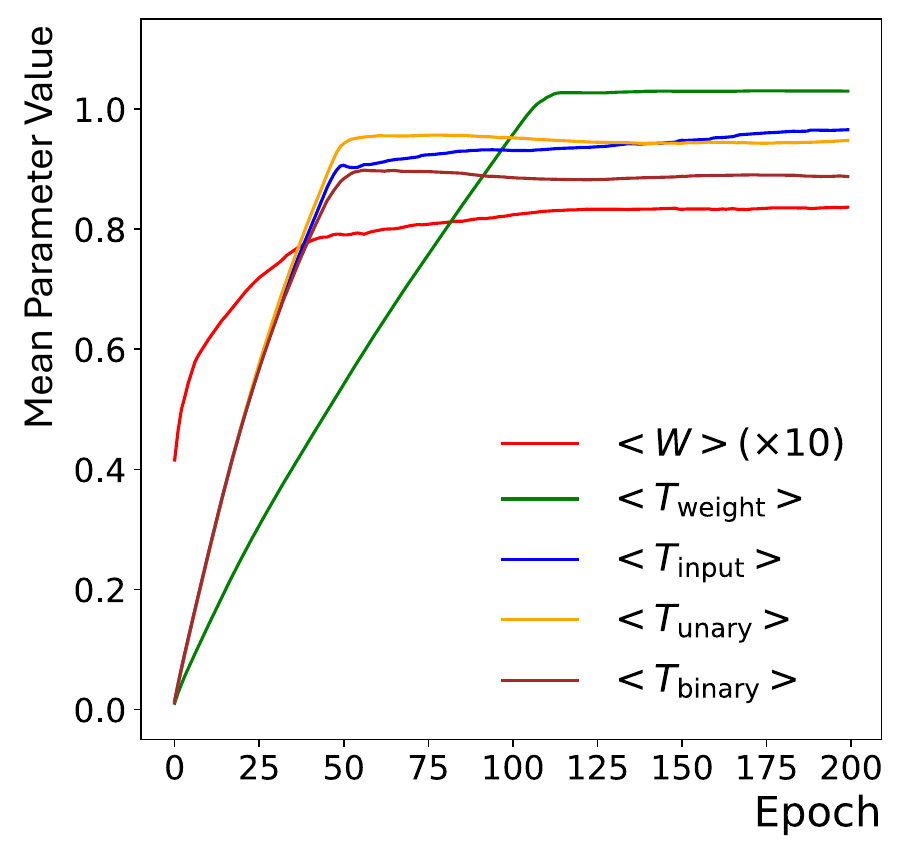}}%
\label{fig:training_lhc_2}
\hfil
\subfloat[]{\includegraphics[width=0.32\textwidth]{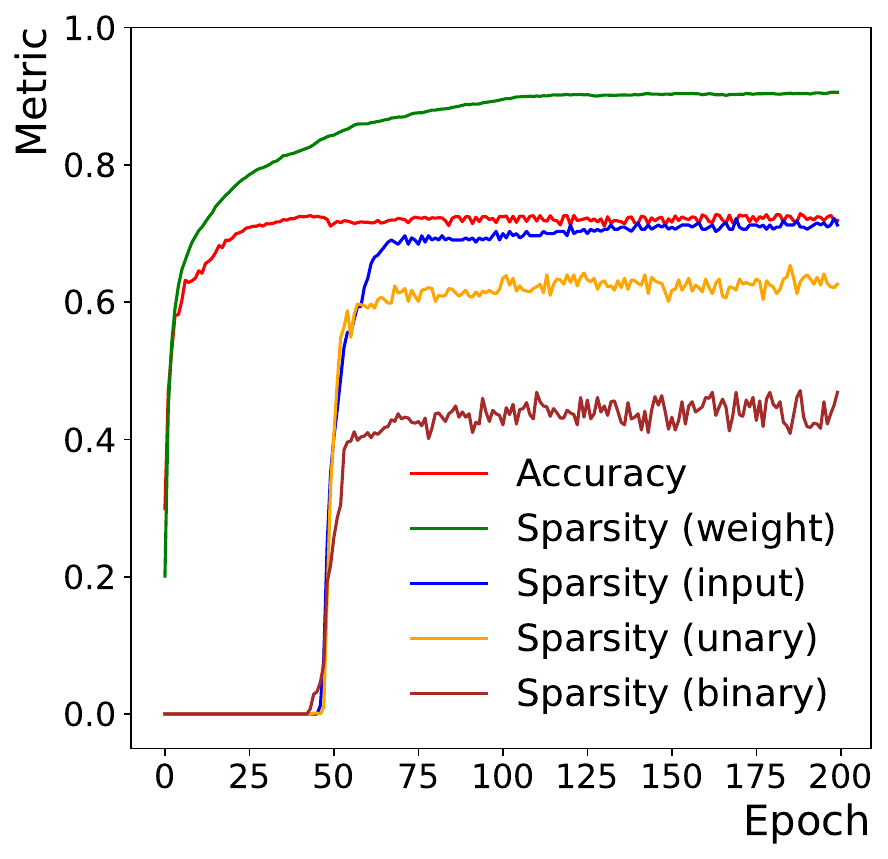}}%
\label{fig:training_lhc_3}
\caption{We demonstrate the training performance of $\tt{SymbolNet}$ on the LHC jet tagging dataset. This $\tt{SymbolNet}$ model consists of two symbolic layers, each containing $u=v=20$ operations. It is trained with a batch size of 1024 for 200 epochs. The $\tt{Adam}$ optimizer \cite{kingma2017adam} is used with a learning rate of 0.0015. The target sparsity ratios are set to $\alpha_{\text{weight}}=0.9$, $\alpha_{\text{input}}=0.75$, $\alpha_{\text{unary}}=0.6$, and $\alpha_{\text{binary}}=0.4$. (a): The individual loss terms in Eq. \ref{eq:loss2} as functions of the epoch. (b): The mean trainable parameters as functions of the epoch. (c): The accuracy and sparsity ratios as functions of the epoch.}
\label{fig:training_lhc}
\end{figure*}

\begin{table*}[!t]
\caption{An example of a compact symbolic model with {\bf{a mean complexity of 16 and an overall accuracy of 71\%}} learned by $\tt{SymbolNet}$ on the LHC jet tagging dataset. Constants are rounded to 2 significant figures for the purpose of display.}
\label{tab:exp_lhc}
\centering
\resizebox{\textwidth}{!}{
\begin{tabular}{|c|l|c|c|}\hline
    \textbf{Class} & \textbf{Expression (symbolic model for LHC jet tagging)} & \textbf{Complexity} & \textbf{AUC} \\ \hline
    $g$ & $-0.041\text{Multiplicity}\bm{\times} C_{1}^{\beta=1} + 0.53\bm{\tanh(}0.6\text{Multiplicity} - 0.38C_{1}^{\beta=1}\bm{)} + 0.24$ & 16 & 0.885 \\ \hline
    $q$ & $0.073\text{Multiplicity}\bm{\times} C_{1}^{\beta=1} - 0.38\bm{\tanh(}0.63m_{\text{mMDT}}\bm{)} + 0.15$ & 12 & 0.827 \\ \hline
    $t$ & $0.2\bm{\sin(}1.2\text{Multiplicity}\bm{)} + 0.43\bm{\sin(}0.49C_{1}^{\beta=2}\bm{)} - 0.2\bm{\tanh(}0.6\text{Multiplicity} - 0.38C_{1}^{\beta=1}\bm{)} + 0.24$ & 24 & 0.915 \\ \hline
    $W$ & $-0.099\bm{\sin(}0.73\text{Multiplicity}\bm{)} + 0.84\bm{\exp(}-46.0\bm{(}m_{\text{mMDT}} + 0.14C_{1}^{\beta=1} + 0.27C_{1}^{\beta=2}\bm{)}^{\bm{2}}\bm{)} + 0.044$ & 23 & 0.894 \\ \hline
    $Z$ & $0.43\bm{\exp(}-6.9\bm{(}C_{1}^{\beta=2}\bm{)}^{\bm{2}}\bm{)}$ & 8 & 0.851 \\ \hline
    \end{tabular}%
    }
\end{table*}

To demonstrate the effectiveness of our adaptive dynamic pruning framework, we conducted a trial using $\tt{SymbolNet}$ with two symbolic layers, each containing $u=v=20$ operators, and set the target sparsity ratios for model weights, inputs, unary operators, and binary operators to $\alpha_{\text{weight}}=0.9$, $\alpha_{\text{input}}=0.75$, $\alpha_{\text{unary}}=0.6$, and $\alpha_{\text{binary}}=0.4$, respectively.
The training curves for this trial are shown in Fig. \ref{fig:training_lhc}.

As seen in Fig. \ref{fig:training_lhc}a, the training loss ($L_{\text{MSE}}$) and the other sparsity terms ($L_{\text{sparse}}$) steadily decrease until around epoch 50.
At this point, the trainable thresholds for the input, unary operator, and binary operator all approach 1 on average, as shown in Fig. \ref{fig:training_lhc}b.
Consequently, many nodes begin to be pruned as their threshold values reach 1.
This pruning is reflected in Fig. \ref{fig:training_lhc}c, where the sparsity ratios increase sharply, with the ratios for unary and binary operators reaching their target values, causing their losses to drop steeply toward zero.

However, the sparsity ratios for both model weights and inputs remain below their target values, so their losses do not yet reach zero.
A small kink is observed in $L_{\text{MSE}}$ (or accuracy) around the same epoch, caused by the steep increase in sparsity ratios.
Despite this, the training process adjusts dynamically, and the training loss continues to decrease as the sparsity ratios steadily increase.
The sparsity ratio for model weights reaches its target value around epoch 100, resulting in a steep drop in its loss at that point.
By the end of the training, the input sparsity ratio reaches around 70\%, slightly below its target of 75\%, so its loss does not fully vanish.
Throughout the training, the sparsity ratios converge toward their target values, with regression and sparsity being optimized simultaneously in this dynamic process.

\begin{figure}[!t]
\centering
\subfloat[]{\includegraphics[width=0.48\textwidth]{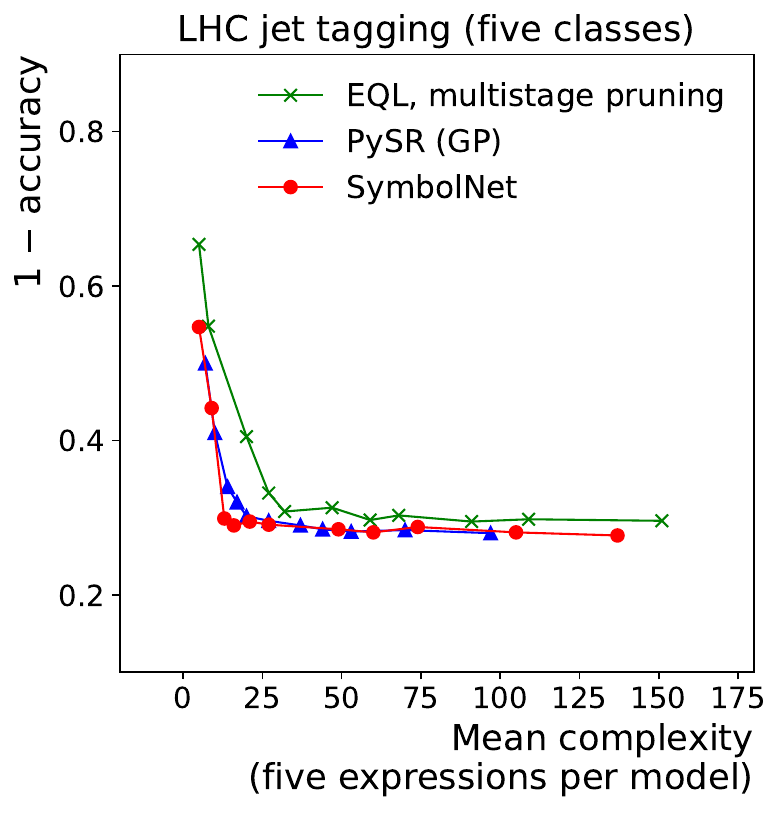}%
\label{fig:lhc_a}}
\hfil
\subfloat[]{\includegraphics[width=0.5\textwidth]{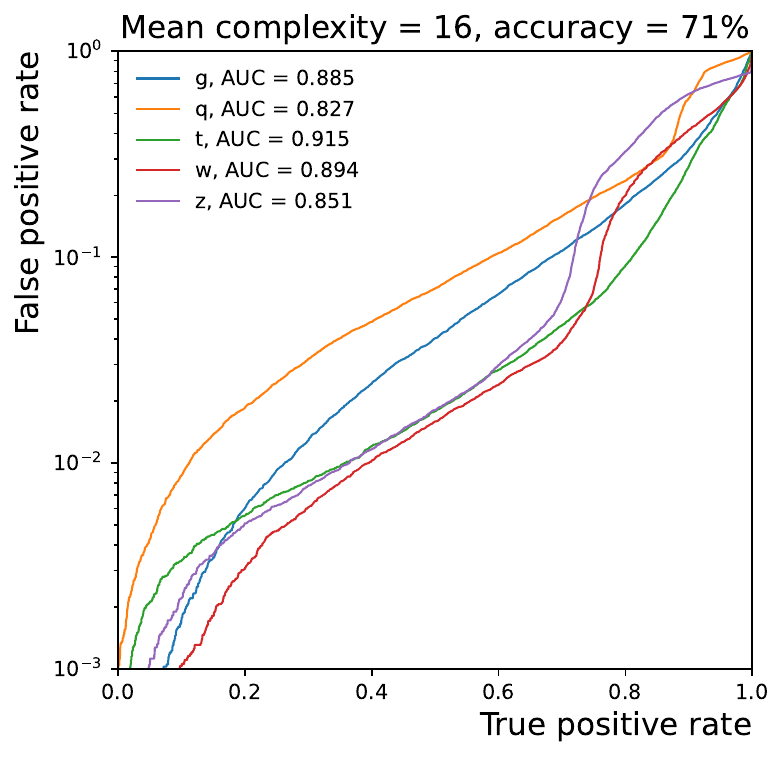}%
\label{fig:lhc_b}}
\caption{Performance of $\tt{SymbolNet}$ on the LHC jet tagging dataset. (a): The Pareto front, each with models selected from the 40 trials, illustrates the trade-off between accuracy and expression complexity. (b): ROC curves for a compact symbolic model with a mean complexity of 16 and an overall accuracy of 71\%, with its expressions tabulated in Tab.~\ref{tab:exp_lhc}.}
\label{fig:lhc}
\end{figure}

Fig. \ref{fig:lhc_a} illustrates the trade-off between accuracy and model complexity, demonstrating that $\tt{SymbolNet}$ outperforms $\tt{EQL}$ in the multistage pruning framework and is comparable to the computationally intensive GP-based algorithm $\tt{PySR}$.

Tab. \ref{tab:exp_lhc} presents an example model learned by $\tt{SymbolNet}$.
This symbolic model is remarkably compact, consisting of only five lines of expressions, with an average complexity of 16, achieving an overall jet-tagging accuracy of 71\%.
Fig. \ref{fig:lhc_b} shows the ROC curve for each of these five expressions.

For comparison, a black-box three-layer NN with $O(10^3)$ parameters achieves an accuracy of 75\%.
However, such a model would require orders of magnitude more resources to compute on an FPGA than a symbolic model with a similar complexity, as studied in \cite{tsoi2023symbolic}.

\subsection{MNIST and binary SVHN}

\begin{figure}[!t]
\centering
\subfloat[]{\includegraphics[width=0.5\textwidth]{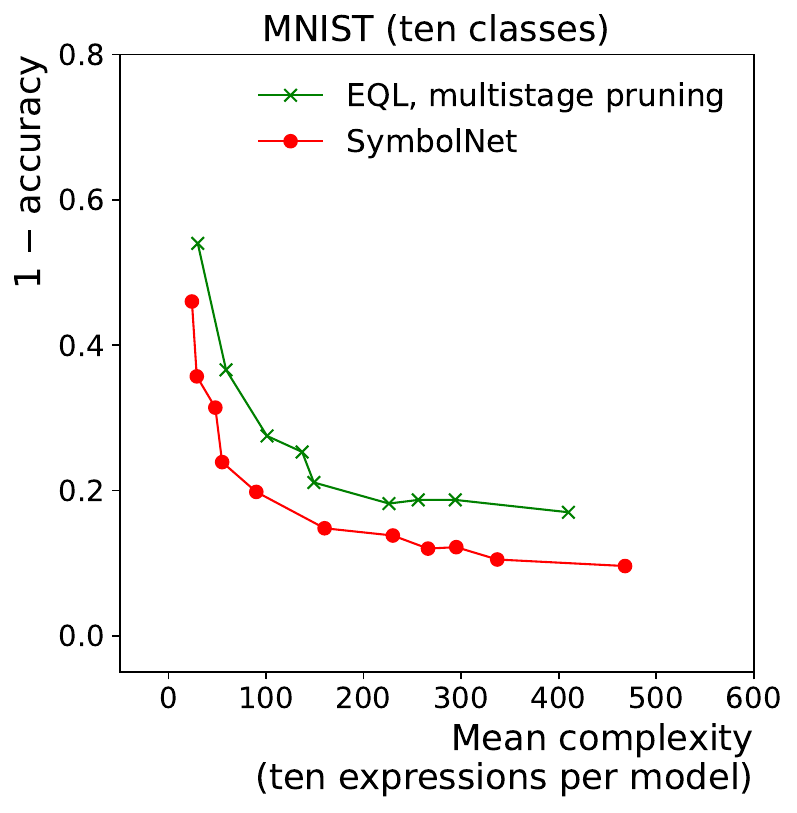}%
\label{fig:mnist}}
\hfil
\subfloat[]{\includegraphics[width=0.5\textwidth]{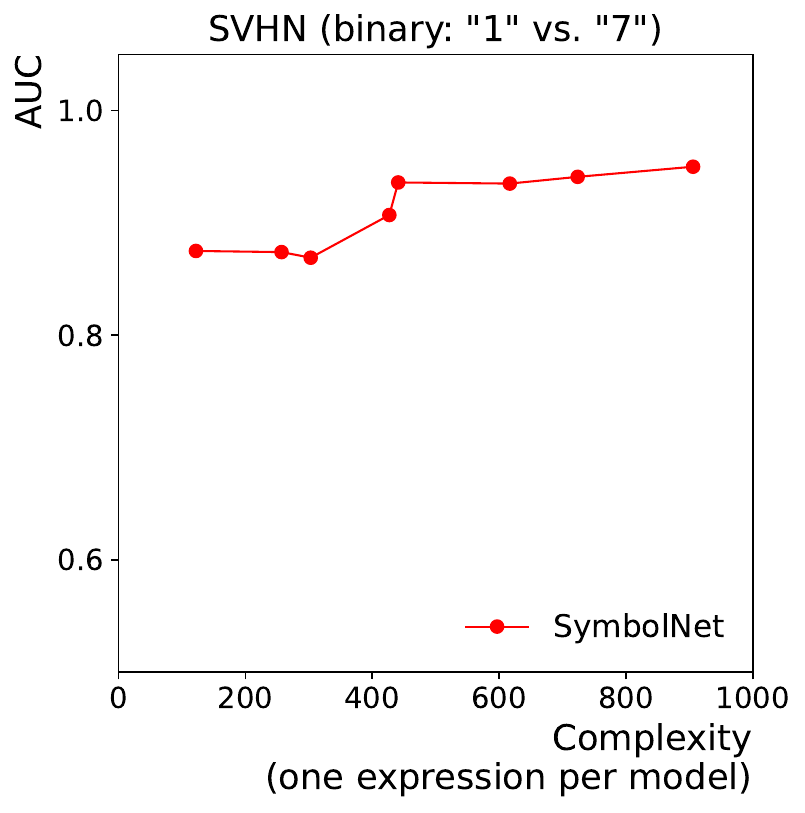}%
\label{fig:svhn}}\\
\subfloat[]{\includegraphics[width=0.48\textwidth]{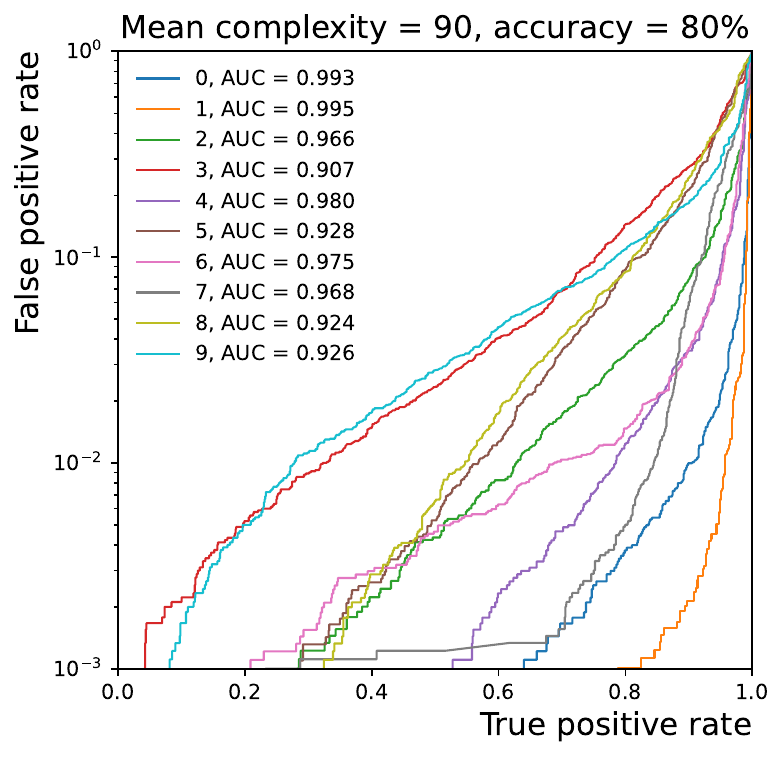}%
\label{fig:roc_mnist}}
\hfil
\subfloat[]{\includegraphics[width=0.5\textwidth]{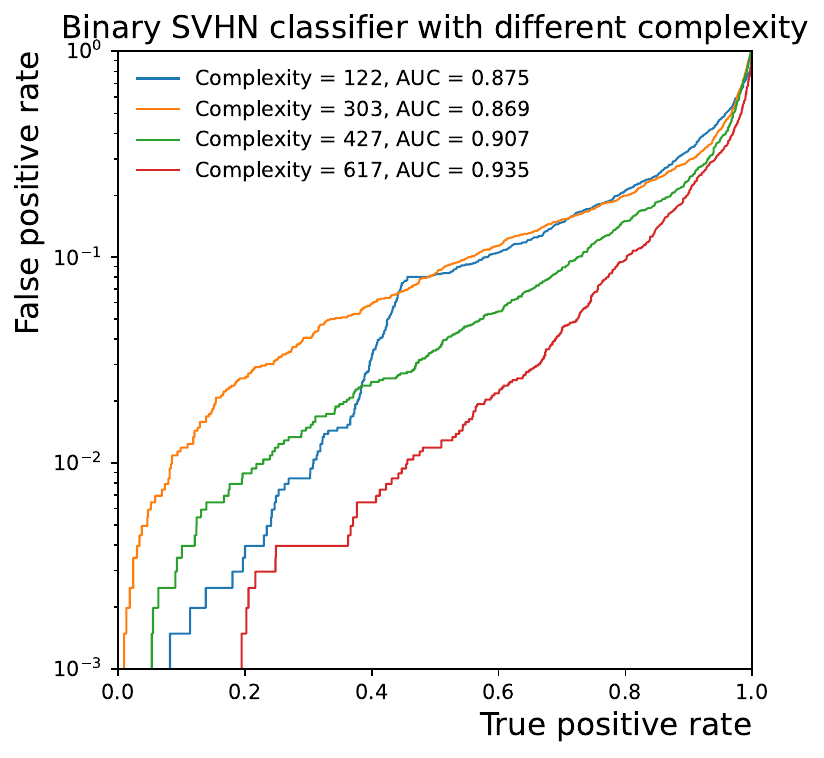}%
\label{fig:roc_svhn}}
\caption{Performance of $\tt{SymbolNet}$ on the MNIST and binary SVHN datasets. (a): The Pareto front, each with models selected from the 40 trials, illustrates the trade-off between accuracy and expression complexity, comparing $\tt{SymbolNet}$ with $\tt{EQL}$ on the MNIST dataset. (b): The ROC AUC, with models selected from 40 trials, is plotted against expression complexity to demonstrate the performance of $\tt{SymbolNet}$ on the binary SVHN dataset. (c): ROC curves for a compact symbolic model with a mean complexity of 90 and an accuracy of 80\% on the MNIST dataset (the expressions of this model are listed in Tab.~\ref{tab:exp_mnist}). (d): ROC curves for four symbolic models with different complexity values on the binary SVHN dataset (the expression for the model with a complexity of 122 is listed in Tab.~\ref{tab:exp_svhn}).}
\label{fig:mnist-svhn}
\end{figure}

\begin{table*}[!t]
\caption{An example of a compact symbolic model with {\bf{a mean complexity of 90 and an overall accuracy of 80\%}}, learned by $\tt{SymbolNet}$ on the MNIST dataset. Constants are rounded to 2 significant figures for the purpose of display.}
\label{tab:exp_mnist}
\centering
\resizebox{\textwidth}{!}{
\begin{tabular}{|c|l|c|c|}\hline
    Class & Expression (symbolic model classifying MNIST digits) & Complexity & AUC \\ \hline
    $0$ & $0.094\bm{\sin(}0.41x_{374} - 0.53x_{378} + 0.66x_{484}\bm{)} + 0.15\bm{(}-0.3x_{184} - 0.17x_{239} - 0.12x_{269} - 0.27x_{271} - $ & 129 & 0.993 \\
    & $ 0.14x_{318} + 0.72x_{352} - 0.6x_{358} - 0.19x_{374} + 0.55x_{377} - 0.32x_{415} - 0.23x_{456} - 0.26x_{485} - 0.4x_{510} - $ & & \\
    & $0.53x_{627} + 0.25x_{637} - 0.19x_{658} + 0.55x_{711}\bm{)\times (}0.44x_{102} - 0.29x_{156}- 0.41x_{212} - 0.29x_{271} - 0.22x_{302} - $ & & \\
    & $ 0.11x_{371} - 0.5x_{398} - 0.41x_{428} - 0.24x_{430}+ 0.84x_{433} + 0.6x_{436} + 0.11x_{462} + 0.62x_{490} - 0.45x_{509} -  $ & & \\ 
    & $0.066x_{539} - 0.4x_{541} - 0.13x_{568} + 0.22x_{580}- 0.58x_{627} - 0.25x_{658}\bm{)}$ & & \\ \hline

    $1$ & $\bm{\exp(}-26.0\bm{(}0.11x_{102} + 0.056x_{158} + 0.21x_{176} + 0.08x_{178} + 0.093x_{182} + 0.93x_{205} + 0.11x_{212} + 0.15x_{235} + $ & 91 & 0.995 \\
    & $0.27x_{248} - 0.033x_{267} + 0.067x_{271} + 0.24x_{302} - 0.063x_{323} + 0.095x_{327} - 0.067x_{350} - 0.12x_{378} + $ & & \\
    & $0.18x_{430} + x_{438} - 0.067x_{462} - 0.092x_{489} + 0.18x_{510} - 0.024x_{568} + 0.12x_{580} + 0.23x_{637} + 0.24x_{711} + $ & & \\
    & $0.13x_{713} + 0.24x_{715} + 0.27x_{96} + 0.28\bm{)^2)}$ & & \\ \hline

    $2$ & $0.54\bm{\sin(}0.59x_{124} + 0.35x_{156} - 0.39x_{318} - 0.41x_{350} - 0.46x_{371} - 0.41x_{374} - 0.6x_{415} + 0.18x_{431} + $ & 58 & 0.966 \\
    & $ 0.14x_{465} + 1.1x_{473} + 0.7x_{509} + 0.38x_{515} + 0.88x_{528} + 0.38x_{554} + 0.77x_{611} + 0.39x_{637} + 0.1x_{99} - $ & & \\
    & $ 0.8\bm{)} + 0.53$ & & \\ \hline

    $3$ & $-0.042x_{158} + 0.062x_{178} - 0.039x_{235} - 0.12x_{291} - 0.063x_{316} + 0.045x_{318} + 0.061x_{404} - 0.066x_{458} + $ & 56 & 0.907 \\
    & $0.032x_{485} - 0.1x_{487} - 0.074x_{489} - 0.12x_{490} + 0.038x_{515} + 0.036x_{517} - 0.06x_{541} + 0.36x_{563} -  $ & & \\
    & $0.043x_{572} + 0.048x_{611} + 0.28$ & & \\ \hline

    $4$ & $0.76\bm{\exp(}-4.7\bm{(}0.47x_{124} + 0.42x_{126} + 0.49x_{128} + 0.14x_{176} + 0.28x_{182} + 0.44x_{184} + 0.17x_{212} + x_{239} + $ & 141 & 0.980 \\
    & $0.88x_{267} + 0.81x_{322} + 0.43x_{323} + 0.33x_{350} + 0.4x_{543} + 0.3x_{554} + 0.5x_{568} + 0.35x_{623}\bm{)^2)} - 0.082\times$ & & \\
    & $\bm{(}-0.2x_{124} - 0.34x_{182} + 0.39x_{429} - 0.69x_{568} - 0.66x_{713} + 0.68\bm{)\times(}1.4x_{102} + 0.58x_{182} + 0.75x_{208} + $ & & \\
    & $0.51x_{215} + 0.29x_{235} + 0.47x_{322} - 0.53x_{323} - 0.7x_{325} + 0.23x_{355} + 0.53x_{358} - 1.4x_{374} - 1.5x_{398} - $ & & \\ 
    & $0.63x_{431} - 1.5x_{456} - 0.68x_{462} - 1.1x_{465} + 0.48x_{541} + 0.83x_{568} + 4.9x_{66} + 1.3x_{71} + 1.3x_{713} + 1.4x_{96}\bm{)} $ & & \\ \hline

    $5$ & $\bm{\exp(}-2.4\bm{(}-0.15x_{124} + 0.13x_{158} + 0.59x_{190} + 0.98x_{248} - 0.13x_{267} - 0.35x_{323} - 0.68x_{325} - x_{327} - $ & 79 & 0.928 \\
    & $ 0.78x_{355} + 0.17x_{404} - 0.5x_{456} - 0.19x_{490} - 0.41x_{510} - 0.6x_{515} + 0.15x_{568} - 0.63\bm{)^2)} -0.012x_{128} - $ & & \\
    & $ 0.12x_{358} + 0.03x_{371} + 0.069x_{374} - 0.031x_{436} - 0.019x_{485} + 0.042x_{580} + 0.026x_{623} $ & & \\ \hline

    $6$ & $0.21x_{102} + 0.3x_{103} + 0.42x_{107} - 0.054x_{215} - 0.057x_{217} - 0.093x_{269} - 0.065x_{271} - 0.068x_{302} - $ & 89 & 0.975 \\
    & $ 0.08x_{322} + 0.068x_{358} + 0.04x_{374} + 0.12x_{414} + 0.021x_{431} + 0.069x_{485} - 0.063x_{489} - 0.078x_{510} +$ & & \\
    & $ 0.081x_{515} + 0.047x_{543} - 0.056x_{568} + 0.065x_{572} - 0.05x_{580} + 0.35x_{64} + 0.43x_{66} + 0.22x_{68} + 0.34x_{69} + $ & & \\
    & $ 0.29x_{71} + 0.35x_{73} + 0.56x_{78} + 0.18x_{99} + 0.1$ & & \\ \hline

    $7$ & $0.98\bm{\exp(}-3.1\bm{(}-x_{124} - 0.61x_{126} - 0.81x_{128} - 0.97x_{156} - 0.24x_{184} - 0.23x_{350} + 0.073x_{355} - 0.28x_{376} - $ & 68 & 0.968 \\
    & $ 0.13x_{377} - 0.62x_{378} - 0.72x_{404} - 0.6x_{415} - 0.62x_{431} - 0.092x_{433} - 0.43x_{458} - 0.87x_{485} - 0.94x_{539} - $ & & \\
    & $ 0.27x_{541} - 0.84x_{581} - 0.37x_{623}\bm{)^2)}$ & & \\ \hline

    $8$ & $- 0.68\bm{\sin(}0.16x_{156} - 0.35x_{176} + 0.43x_{302} + 0.19x_{318} + 0.23x_{327} + 0.41x_{376} - 0.2x_{414} - 0.4x_{428} + $ & 55 & 0.924 \\
    & $ 0.46x_{433} - 0.33x_{467} + 0.27x_{487} + 0.3x_{515} - 0.34x_{528} + 0.25x_{541} + 0.58x_{658} + 0.43x_{689} + 1.1\bm{)} + 0.64$ & & \\ \hline

    $9$ & $- 0.051\bm{\sin(}0.59x_{124} + 0.35x_{156} - 0.39x_{318} - 0.41x_{350} - 0.46x_{371} - 0.41x_{374} - 0.6x_{415} + 0.18x_{431} + $ & 136 & 0.926 \\
    & $ 0.14x_{465} + 1.1x_{473} + 0.7x_{509} + 0.38x_{515} + 0.88x_{528} + 0.38x_{554} + 0.77x_{611} + 0.39x_{637} + 0.1x_{99} - 0.8\bm{)} - $ & & \\
    & $ 0.054x_{126} - 0.066x_{158} - 0.082x_{190} - 0.11x_{205} + 0.059x_{208} + 0.016x_{215} - 0.039x_{217} - 0.0092x_{235} - $ & & \\
    & $ 0.11x_{248} - 0.047x_{271} + 0.093x_{316} - 0.04x_{322} + 0.069x_{327} + 0.07x_{352} - 0.059x_{414} + 0.069x_{429} +  $ & & \\
    & $ 0.038x_{431} + 0.048x_{436} - 0.057x_{467} - 0.044x_{517} - 0.067x_{541} + 0.065x_{637} - 0.06x_{658} + 0.11x_{711} +  $ & & \\
    & $ 0.1x_{713} + 0.16x_{715} + 0.029$ & & \\ \hline
    
    \end{tabular}%
    }
\end{table*}

\begin{table*}[!t]
\caption{An example of a compact symbolic model with {\bf{a complexity of 122 and an ROC AUC of 0.875}}, learned by $\tt{SymbolNet}$ on the binary SVHN dataset (classes `1' and `7'). Constants are rounded to 2 significant figures for the purpose of display.}
\label{tab:exp_svhn}
\centering
\resizebox{\textwidth}{!}{
\begin{tabular}{|l|c|c|}\hline
    Expression (symbolic model classifying SVHN digits in a binary setting: `1' vs. `7') & Complexity & AUC \\ \hline
    $0.58\bm{\exp(}-4.7\bm{(}0.35x_{1191} + 0.2x_{1282} + 0.29x_{1285} + 0.53x_{1384} + 0.3x_{1566} + 0.35x_{1788} - 0.56x_{2156} + 0.38x_{2179} + $ & 122 & 0.875 \\
    $ 0.51x_{2460} + 0.22x_{2470} + 0.33x_{2746} - 0.6x_{429} - 0.26x_{612} - 0.45x_{628} - 0.32x_{637} - 0.33x_{732} - 0.26x_{733} - 0.28x_{813} - $ & & \\
    $ 0.15x_{913} - 1\bm{)^2)} + 0.61\bm{\cos(}1.4x_{1282} - 1.4x_{1298} - 1.6x_{1486} + 1.7x_{1863} - 1.2x_{2357} - 0.53x_{2460} + 0.79x_{2609} + $ & & \\
    $0.79x_{3046} - 0.94x_{485} + 0.71x_{511} - 1.4x_{527} + 1.9x_{617} + 0.76x_{637} - 1.8x_{720} + 1.8x_{824} + 0.68x_{831} - 0.99x_{913}\bm{)}$ & & \\ \hline
    \end{tabular}%
    }
\end{table*}

We analyze the MNIST dataset by considering all ten classes, with each symbolic model generating ten expressions, each corresponding to one of the ten classes.
Fig. \ref{fig:mnist} shows a Pareto front generated by $\tt{SymbolNet}$ on the MNIST dataset, outperforming $\tt{EQL}$ in the multistage pruning framework.
For instance, $\tt{SymbolNet}$ can achieve an overall accuracy of 90\% with a mean complexity of around 300.
Tab. \ref{tab:exp_mnist} presents the ten expressions of a symbolic model learned by $\tt{SymbolNet}$, which has a mean complexity of 90 and an overall accuracy of 80\%.
The ROC curves for these expressions are shown in Fig. \ref{fig:roc_mnist}.
This example demonstrates the power of symbolic models learned by $\tt{SymbolNet}$---compact enough to be fully visualized within a table yet capable of making reasonable predictions.

For the SVHN dataset in the binary setting, each symbolic model generates a single expression.
We observed that $\tt{EQL}$ struggled to converge on this dataset, producing either overly complex expressions with reasonable accuracy or models that were too sparse to make meaningful predictions.
In contrast, $\tt{SymbolNet}$ was able to scale to this high-dimensional dataset and generated compact expressions with reasonable predictive accuracy.
Fig. \ref{fig:svhn} shows the ROC AUC as a function of expression complexity for models learned by $\tt{SymbolNet}$, while Fig. \ref{fig:roc_svhn} shows the ROC curves for four symbolic models with different complexity values.
Tab. \ref{tab:exp_svhn} lists the expression of a symbolic model with a complexity of 122 and an ROC AUC of 0.875.
Remarkably, even such a single-line expression can provide decent predictions in classifying SVHN digits from noisy real-world scenes, as depicted in Fig. \ref{fig:input_svhn}.

\subsection{FPGA resource utilization and latency}

\begin{table*}[!t]
\caption{Resource utilization and latency on an FPGA for quantized and pruned (QP) NNs and symbolic expressions learned by $\tt{SymbolNet}$. The model size is expressed in terms of the number of neurons per hidden layer for DNNs and the number of filters for CNNs, where, for example, $(16)_{3}$ indicates 16 filters with a kernel size of 3$\times$3. The initiation interval (II) is quoted in clock cycles. The numbers in parentheses indicate the percentage of total available resource utilization. The relative accuracy and ROC AUC are evaluated with respect to the same DNN/CNN implemented in floating-point precision and without pruning.}
\label{tab:fpga}
\centering
\resizebox{\textwidth}{!}{
\begin{tabular}{|l|c|c|c|c|c|c|c|c|c|}
    \hline \multicolumn{10}{|c|}{\bf{LHC jet tagging (five classes)}} \\ \hline
    
    & Model size (input dim. $=$ 16) & Precision & BRAMs & DSPs & FFs & LUTs & II & Latency & Rel. acc. \\ \hline
    
    QP DNN & [64, 32, 32, 5], \bf{90\% pruned} & $\langle$6, 0$\rangle$ & 4 (0.1\%) & 28 (0.4\%) & 2739 (0.1\%) & 7691 (0.7\%) & 1 & 55 ns & 94.7\% \\ \hline
    
    SR & \bf{Mean complexity of the five expr. $=$ 18} & $\langle$12, 8$\rangle$ & 0 (0\%) & 3 (0\%) & 109 (0\%) & 177 (0\%) & 1 & \bf{10 ns} & 93.3\% \\ \hline
    
    \multicolumn{10}{|c|}{\bf{MNIST (ten classes)}} \\ \hline
    & Model size (input dim. $=$ 28$\times$28$\times$1) & Precision & BRAMs & DSPs & FFs & LUTs & II & Latency & Rel. acc. \\ \hline %
    
    QP CNN & [(16, 16, 24$)_{\text{3}}$, 42, 64, 10], \bf{92\% pruned} & $\langle$6, 0$\rangle$ & 66 (1.5\%) & 216 (3.2\%) & 18379 (0.8\%) & 29417 (2.5\%) & 788 & 4.0 $\mu$s & 86.8\% \\ \hline 
    
    SR & \bf{Mean complexity of the ten expr. $=$ 133} & $\langle$18, 10$\rangle$ & 0 (0\%) & 160 (2.3\%) & 6424 (0.3\%) & 7592 (0.6\%) & 1 & \bf{125 ns} & 85.3\% \\ \hline
    
    \multicolumn{10}{|c|}{\bf{SVHN (binary `1' vs. `7')}} \\ \hline
    & Model size (input dim. $=$ 32$\times$32$\times$3) & Precision & BRAMs & DSPs & FFs & LUTs & II & Latency & Rel. AUC \\ \hline
    
    QP CNN & [(16, 16, 24$)_{\text{3}}$, 42, 64, 1], \bf{92\% pruned} & $\langle$6, 0$\rangle$ & 62 (1.4\%) & 77 (1.1\%) & 16286 (0.7\%) & 27407 (2.3\%) & 1029 & 5.2 $\mu$s & 94.0\% \\ \hline 
    
    SR & \bf{Complexity $=$ 311} & $\langle$10, 4$\rangle$ & 0 (0\%) & 38 (0.6\%) & 1945 (0.1\%) & 3029 (0.3\%) & 1 & \bf{195 ns} & 94.5\% \\ \hline
    
    \end{tabular}%
    }
\end{table*}

Tab. \ref{tab:fpga} presents a comparison of resource utilization and latency on an FPGA between symbolic models and NNs with typical compression techniques.
The FPGA resources considered include on-board FPGA memory (BRAMs), digital signal processors (DSPs), flip-flops (FFs), and lookup tables (LUTs), as estimated from the logic synthesis step.
Symbolic models generally consume significantly fewer resources and require much less latency than NNs, even when the NNs are strongly quantized and pruned, while still achieving comparable accuracy across the three datasets.

%% file: 6_conclusion.tex
\section{Discussion}
\label{sec-conclusion}

We have introduced $\tt{SymbolNet}$, a neural network-based approach to symbolic regression for model compression, featuring a novel pruning framework designed to generate compact expressions capable of fitting high-dimensional data.
Our method aims to address the latency and bandwidth bottlenecks encountered in modern high-energy physics experiments, such as those at the CERN LHC, where datasets are typically high-dimensional and do not have ground-truth equations.
By leveraging the efficiency of compact symbolic representations, which can be implemented on custom hardware such as FPGAs, we further extend the applicability of symbolic regression to higher-dimensional data, as commonly found in LHC experiments.

Most existing symbolic regression methods, whether based on genetic programming or deep learning, primarily focus on datasets with input dimensions less than $\mathcal{O}(10)$ and cannot scale efficiently beyond.
In genetic programming approaches, the search algorithms create and evolve equations using a combinatorial strategy, which becomes highly inefficient as the equation search space scales exponentially with its building blocks.
Even with an external feature selector, the GP-based algorithms can still be impractical for datasets with input dimensions beyond $\mathcal{O}(100)$.
On the other hand, while deep learning techniques have demonstrated their ability to handle complex datasets in other domains, they have not been extensively explored for symbolic regression in high-dimensional problems.

Our proposed method, equipped with a novel pruning framework dedicated to symbolic regression, aims to fill this gap.
Its effectiveness has been demonstrated on datasets with input dimensions ranging from $\mathcal{O}(10)$ to $\mathcal{O}(1000)$.
While there is always a trade-off between prediction accuracy and computational resources, this work provides an option to minimize the latter.
Our method is specifically designed to be a model compression technique by leveraging the compact representation of symbolic expressions for low-latency deployment in resource-constrained environments.

However, neural networks currently have certain limitations when used for symbolic regression compared to genetic programming.
For example, it is challenging to create a unified framework that can incorporate mathematical operators that are not differentiable everywhere, such as logarithms and division, without specifically regularizing each of these operators.
While these operators could expand model expressivity, they can introduce singular points in gradient-based optimization.
Potential future extensions include tackling time series or sequence data, integrating domain-specific constraints or prior knowledge, and developing methods to improve the interpretability of generated expressions from high-dimensional datasets.
To fully maximize the potential of efficient hardware deployment of symbolic expressions, further work is needed to optimize the expressions while maintaining numerical precision within fixed bit widths, a process known as quantization-aware training~\cite{Coelho_2021}.
These topics will be addressed in future work.